\PassOptionsToPackage{table}{xcolor}
\documentclass[]{fairmeta}

\usepackage{amsmath,amsfonts,bm}

\def\eqref#1{equation~\ref{#1}}

\def\plaineqref#1{\ref{#1}}

\def\1{\bm{1}}

\DeclareMathAlphabet{\mathsfit}{\encodingdefault}{\sfdefault}{m}{sl}
\SetMathAlphabet{\mathsfit}{bold}{\encodingdefault}{\sfdefault}{bx}{n}

\usepackage[utf8]{inputenc}
\usepackage{url}
\usepackage{amsmath}
\usepackage{amssymb}
\usepackage{amsfonts}
\usepackage{array}
\usepackage{arydshln}
\usepackage{algorithm}
\usepackage{algpseudocode}
\usepackage{enumitem}

\makeatletter
\@ifundefined{theHALG@line}{%
  \newcommand{\theHALG@line}{\thealgorithm.\arabic{ALG@line}}%
}{%
  \renewcommand{\theHALG@line}{\thealgorithm.\arabic{ALG@line}}%
}
\makeatother

\let\cite\citep

\definecolor{thinkcolor}{RGB}{70,130,180}
\definecolor{plancolor}{RGB}{34,139,34}
\definecolor{toolcolor}{RGB}{178,90,20}
\definecolor{answercolor}{RGB}{128,0,128}

\tcbset{
  promptbox/.style={
    breakable,
    colback=gray!5,
    colframe=gray!50,
    fonttitle=\bfseries,
    arc=2mm,
    fontupper=\small,
    fontlower=\small,
    before upper=\sloppy,
    before lower=\sloppy,
    segmentation style={dashed, gray!55},
  }
}

\newtcbox{\rqtag}{
  on line,
  colback=white,
  colframe=gray!60,
  coltext=black,
  boxrule=0.6pt,
  arc=2pt,
  boxsep=0pt,
  left=3pt,
  right=3pt,
  top=1pt,
  bottom=1pt,
  fontupper=\bfseries\small
}

\makeatletter
\newcommand{\appendixsectionentry}[3]{%
  \par\addvspace{0.9em}%
  \noindent
  \hyperref[#2]{\sffamily\bfseries\makebox[2.2em][l]{#1}#3}%
  \hfill
  \hyperref[#2]{\sffamily\bfseries\pageref*{#2}}%
  \par\nobreak\vspace{0.2em}%
}
\newcommand{\appendixsubsectionentry}[3]{%
  \@dottedtocline{2}{2.2em}{2.6em}%
    {\hyperref[#2]{\numberline{#1}#3}}%
    {\hyperref[#2]{\pageref*{#2}}}%
}
\makeatother

\newcommand{\logoimg}[2][0.95cm]{\raisebox{-0.5\height}{\includegraphics[height=#1]{#2}}}
\newcommand{\logorow}{%
  \par\vspace{0.3cm}%
  {\setlength{\parskip}{0pt}\noindent
  \logoimg{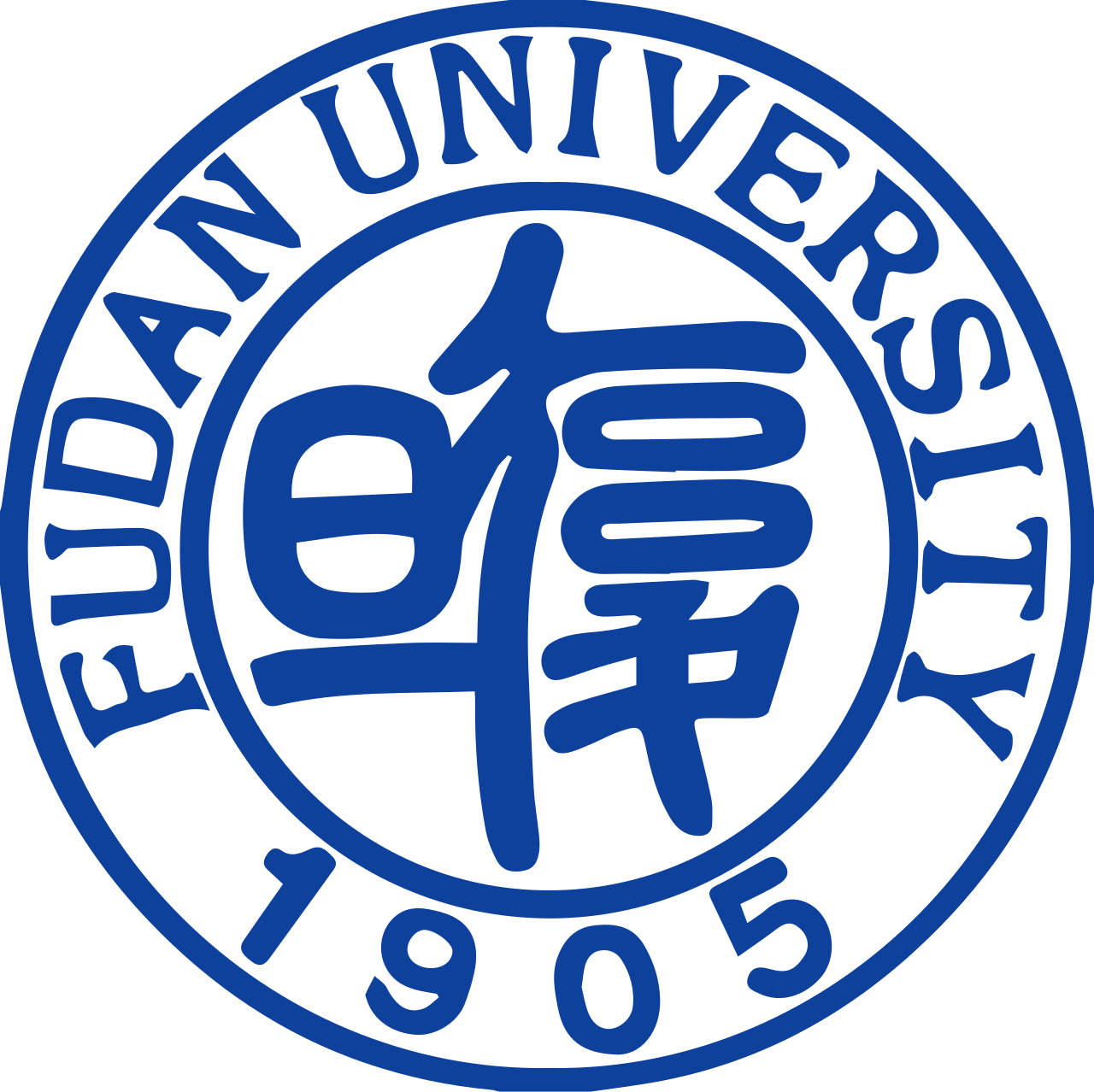}\hspace{0.45cm}\logoimg{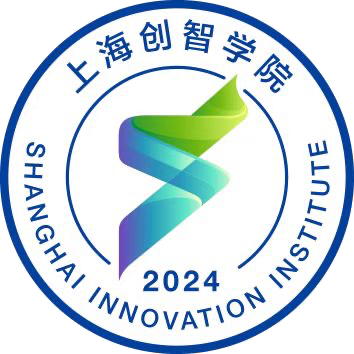}\hspace{0.45cm}\logoimg{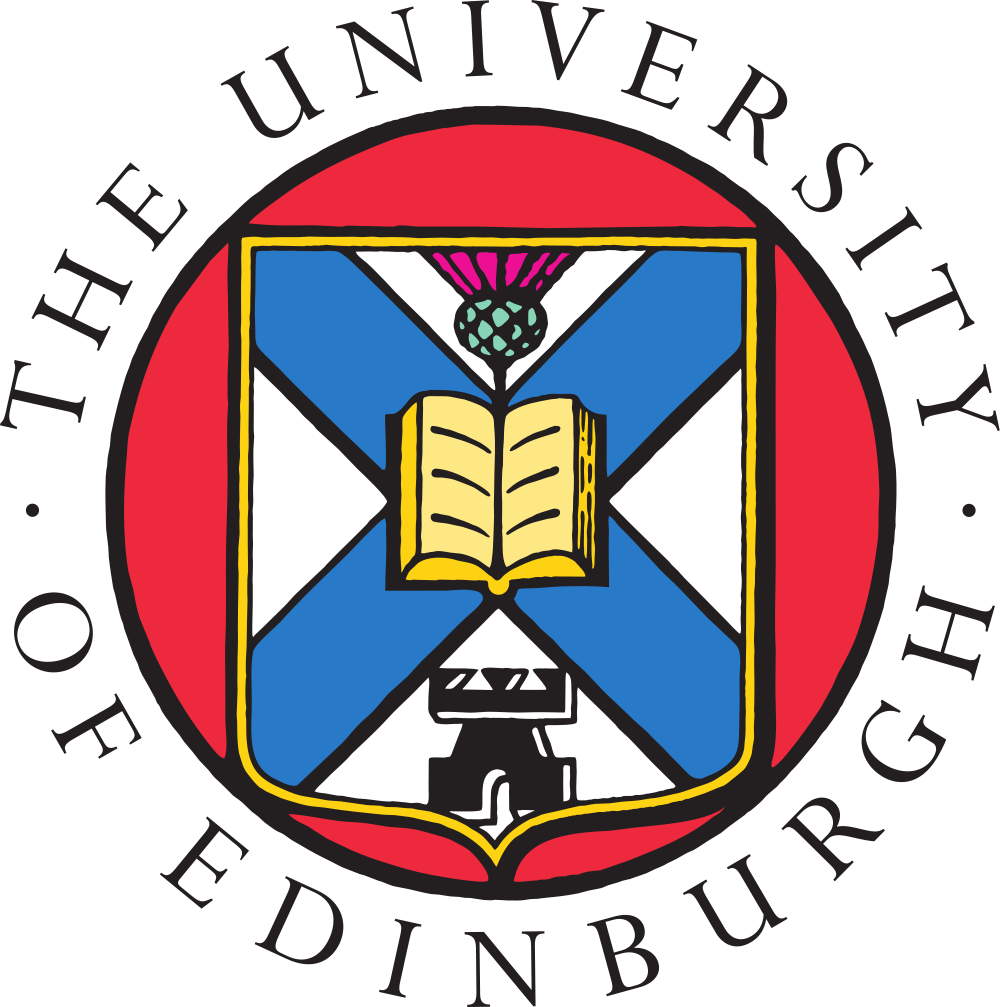}\hspace{0.45cm}%
  \logoimg{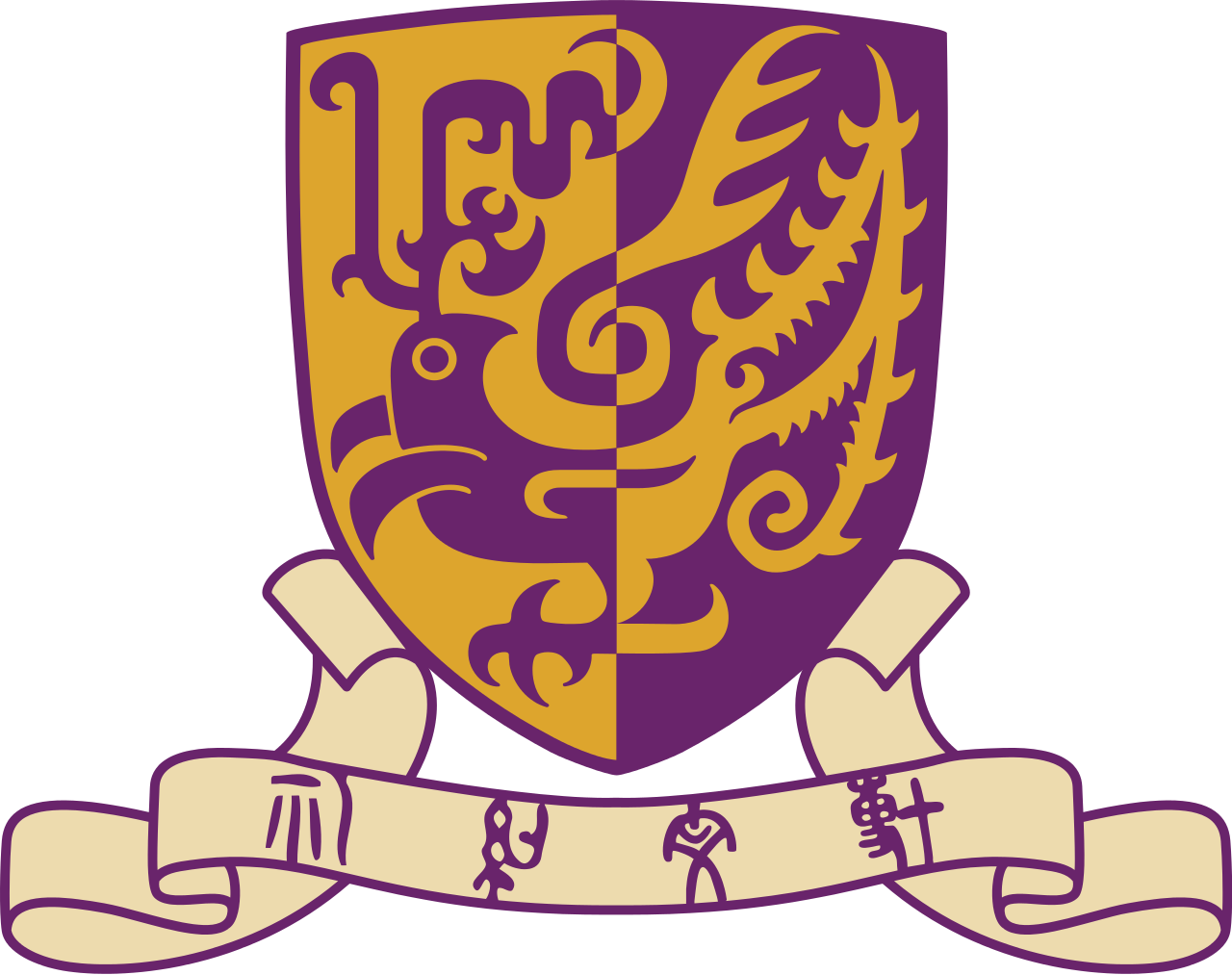}\hspace{0.45cm}\logoimg{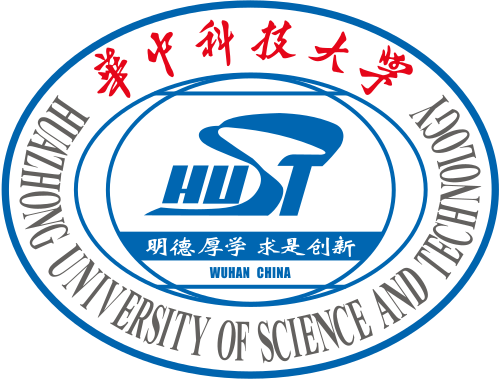}\hspace{0.45cm}\logoimg[0.42cm]{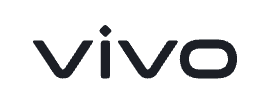}\par}%
}
\makeatletter
\patchcmd{\mymaketitle}{\end{tcolorbox}}{\logorow\end{tcolorbox}}{}{\PatchFailed}
\makeatother

\title{\textsc{ParaAgent}: Reinforcing Parallel Acting in Open-World Tool Environments}

\author[1]{Shengbin Yue}
\author[3]{Hongru Wang}
\author[4]{Siyuan Wang}
\author[6]{Xiaoxin Chen}
\author[5]{Wei Chen}
\author[1,2]{Zhongyu Wei}

\affiliation[1]{Fudan University}
\affiliation[2]{Shanghai Innovation Institute}
\affiliation[3]{University of Edinburgh}
\affiliation[4]{Chinese University of Hong Kong}
\affiliation[5]{Huazhong University of Science and Technology}
\affiliation[6]{Vivo AI}

\abstract{Language model agents are increasingly deployed in open-world tool environments, which require balancing exploring unknown capabilities and exploiting known ones.
Existing methods face a performance--efficiency tradeoff: they either rigidly decouple exploration and execution or interleave them without coordination.
We argue that the key lies not in whether to decouple or interleave them, but in how to coordinate them across granularities.
We introduce \textsc{ParaAct}, a structured parallel-action loop that combines phase-level Exploration~$\rightleftharpoons$~Execution with action-level parallelism.
To learn this loop, \textsc{ParaAgent} combines multi-agent cold-start demonstrations with reinforcement learning under multi-level advantage decoupling, making planning structure explicit and supervising it with step-, phase-, and trajectory-level rewards.
Learning is supported by our \textsc{ToolEnv}, a scalable simulator grounded in 50{,}011 realistic tool interfaces.
On two open-world tool benchmarks, \textsc{ParaAgent}$_{4B}$ achieves the best average success among all baselines, including GPT-4.1 systems, with the largest gains on multi-tool tasks.
Behavioral analyses show that these gains stem from this action organization, highlighting its importance for capable and efficient open-world agents.}

\metadata[Project]{\url{https://github.com/yueshengbin/paraagent}}
\correspondence{Shengbin Yue at \email{sbyue23@m.fudan.edu.cn}, Zhongyu Wei at \email{zywei@fudan.edu.cn}}

\begin{document}

\maketitle

\section{Introduction}
Recent progress in large language models (LLMs) has accelerated the transition from chatbots toward general agentic assistants~\cite{wei2026agentic,guo2025deepseek,zheng2025deepresearcher,cheng2026theory,yue2025synergistic}.
Alongside this, the external tool ecosystem is rapidly expanding in scale and diversity, spanning APIs~\mbox{\cite{qin2024toolllm}}, MCP~\cite{hou2025model}, and Skills~\cite{zhang2025equipping}.
This is propelling LLM agents into \textbf{open-world tool environments}, where task constraints and tool utility often emerge progressively through interaction rather than specified in advance~\cite{yao2023react}.
Effective operation therefore requires balancing \emph{exploration} of unfamiliar capabilities with \emph{exploitation} of acquired information.

\begin{figure}[ht]
    \centering
\includegraphics[width=\linewidth]{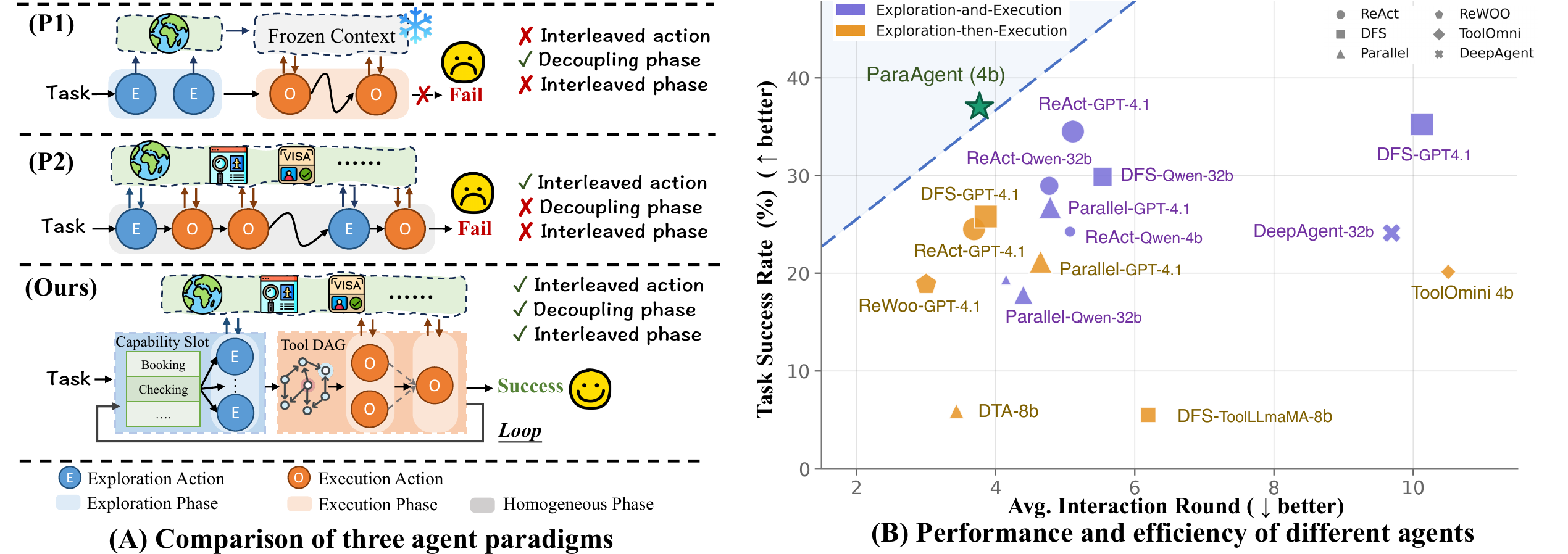}\par
\vspace{-2mm}
\caption{\textbf{Agent paradigms and their performance--efficiency tradeoff.} (A) \emph{Exploration-then-Execution} (P1) rigidly decouples exploration and execution, while native \emph{Exploration-and-Execution} (P2) interleaves them without coordination. 
Ours separate phase-level adaptation from action-level coordination.
(B)~Task success (Pass@1) versus average interaction rounds per task for different agents on ToolBench and API-Bank.}
\label{fig:intro}
\vspace{-4mm}
\end{figure}

Existing approaches broadly follow two paradigms, as shown in Figure~\ref{fig:intro}.
\emph{(1) Exploration-then-Execution}~\cite{li2023api,qu2024towards,kachuee2025improving,huang2026toolomni} first explores a potential tool space, and then acts within this fixed context.
Such decoupling enables tailored phase-wise optimization, facilitating more effective action strategies. However, early exploration errors or emergent requirements may remain unaddressed and ultimately limit task success.
\emph{(2) Exploration-and-Execution} can flexibly perform two types of actions, but existing methods~\cite{qin2025meta,li2026deepagent} naively interleave them as homogeneous actions in a flat sequence.
Without explicit phase structure or coordination, this flexibility relies on frequent serial interleaving, accumulating irrelevant context over time.
This exposes a clear \textbf{performance-efficiency tradeoff}: the former is efficient but brittle, whereas the latter is adaptive but interaction heavy.

The key is not whether to separate or interleave exploration and execution, but how to coordinate them across granularities, \emph{phase-level adaptation} with \emph{action-level parallelism}.
Two challenges arise in
\textbf{(i) Structure}: exploration is \emph{epistemic}, reducing uncertainty about required capabilities, whereas execution is \emph{operational}, coordinating under dependency.
This asymmetry requires specialized parallel strategies and phase transitions when capability gaps emerge.
\textbf{(ii) Learning}: the loop relies on structural decisions that in-context or imitation learning can initialize, but robustly optimizing them requires interaction~\cite{zhang2025landscape}.
It further requires rewarding how actions are organized,
as such decisions are unobservable in individual actions, with effects emerging only phases later.

To address \textbf{(i)}, we introduce \textsc{\textbf{ParaAct}}, a \textbf{structured parallel-action loop} framework that integrates verbal reasoning, explicit planning, and parallel actions.
At the phase level, it drives an \emph{Exploration~$\rightleftharpoons$~Execution} phase loop to expand capabilities as requirements evolve.
At the action level,
two primitives match the asymmetry:
\emph{Exploration parallelism} infers required capability slots and issues parallel queries to fill them.
\emph{Execution parallelism} selects acquired tools and organizes their dependencies as a directed acyclic graph (DAG) for parallel execution.
Together, this framework supports phase-level adaptation while coordinating parallel actions within each phase.

To address \textbf{(ii)}, we further introduce \textsc{\textbf{ParaAgent}}, an \textbf{agentic learning} framework that learns the structured parallel-action loop in open-world tool environments.
\textsc{ParaAgent} first bootstraps the loop through a cold-start stage using a state-sharing multi-agent system: a \emph{planner} controls phase transitions based on a \emph{critic}'s evidence-gap assessment, while a \emph{retriever} and \emph{executor} implement the two parallel primitives.
Although these demonstrations initialize the loop, they cannot ensure robust structural decisions under evolving interactions, while standard GRPO~\cite{shao2024deepseekmath} relies on outcome rewards that fail to capture action organization and dependencies.
Hence, we introduce a reinforcement learning stage with multi-level advantage decoupling.
Our key idea is to make implicit planning structure explicit through phase-level plans, making structural decisions observable and supervisable at the appropriate granularity.
It comprises two key components:
\textbf{ (1) Hierarchical rewards.}
\emph{Step rewards} assess compliance with the plan protocol and formatting requirements; \emph{phase rewards} evaluate planning and execution against the subgraph induced by declared tools in our global tool-dependency graph; and \emph{trajectory rewards} assess final task outcomes via LLM- or rule-based evaluation together with reference-tool hit rate.
Because phase rewards rely on self-generated plans, positive phase rewards are further gated by the final task outcome.
\textbf{(2) Decoupled optimization.} Since naively aggregating heterogeneous rewards may obscure individual learning signals, we independently normalize three reward components, thereby balancing supervision across levels.
The resulting learning signal links action organization to task outcomes, guiding the policy to refine its structural decisions through interaction.

To support this interactive training, we build \textsc{ToolEnv}, a scalable tool environment grounded in real-world tool interactions.
We curate a large-scale corpus of \emph{request--response pairs} and, following prior tool-world-model approaches~\cite{guo2024stabletoolbench,li2025word}, fine-tune LLMs on these pairs and combine them with caching and rule-based methods into a realistic, scalable testbed.

Extensive experiments on ToolBench and API-Bank compare \textsc{ParaAgent} against baselines spanning diverse action strategies, including \emph{serial}, \emph{tree}, and \emph{parallel} acting.
\textsc{ParaAgent}$_{4B}$ achieves the best average success, outperforming the strongest baseline and even GPT-4.1-based systems, with the largest gains on multi-tool tasks.
Behavior analyses further show that these gains come from better action decisions rather than more interaction, and ablations validate the effectiveness of our two-stage learning and decoupled hierarchical rewards.
Experimental results demonstrate the effectiveness and efficiency of learning our structured parallel-action loop across action strategies and model scales.

\section{Related Work}
\paragraph{Open-World Tool Use.}
Real-world agents increasingly operate in large, heterogeneous, and evolving tool environments, where required capabilities are unknown a priori and must be discovered during interaction~\cite{qin2024toolllm,hou2025model,xu2025toucan,zhou2026externalization,li2023api,jia2026ready,yue2025multi}.
Existing methods broadly follow two paradigms.
\emph{Exploration-then-Execution}~\cite{zhang2023retrieve,patil2024gorilla,xu2024enhancing,kachuee2025improving,qin2024toolllm} retrieves a fixed tool set before execution, with improvements through query rewriting, tool-document augmentation~\cite{zhang2023retrieve,xu2024enhancing,kachuee2025improving,shi2025retrieval}, or dedicated tool retrievers~\cite{qin2024toolllm}.
While enabling stage-specific optimization, fixing exploration before execution limits adaptation to emerging requirements.
\emph{Exploration-and-Execution}~\citep{qin2025meta,li2026deepagent,huang2026toolomni} instead dynamically discovers and invokes tools during reasoning.
Meta-Tool~\citep{qin2025meta} integrates retrieval into tool-use reasoning, ToolOmni~\citep{huang2026toolomni} jointly optimizes proactive retrieval and grounded execution, and DeepAgent~\citep{li2026deepagent} unifies tool discovery and execution within an end-to-end reasoning process.
However, these methods conflate phase-level and action-level coordination, limiting both efficiency and flexibility.
\textsc{ParaAgent} separates phase-level adaptation from action-level coordination.

\paragraph{Agentic Learning.}
Agentic learning~\cite{wei2026agentic,zeng2024agenttuning,qu2025tool,yue2025synergistic,zhang2025landscape} enhances LLMs to autonomously reason and act through environment interaction.
Prompt-based methods~\citep{yao2023react,shinn2023reflexion,xu2023rewoo} enable tool use by interleaving reasoning and actions, but remain bounded by in-context capabilities.
Demonstration fine-tuning~\citep{qin2024toolllm,zeng2024agenttuning,yue2025synergistic,chen2024agent} improves task-specific behavior, yet depends on demonstration coverage and quality.
Reinforcement learning~\citep{shao2024deepseekmath,guo2025deepseek,wang2025ragen,wei2025webagent} instead optimizes task success directly through interaction.
However, existing RL methods face two limitations in open-world settings:
\emph{environment-wise}, current environments focus on closed, sequential tool interactions~\citep{jin2025search,feng2025retool,qian2026toolrl}, lacking realistic simulation of large heterogeneous tool spaces;
\emph{algorithm-wise}, flat outcome rewards provide little supervision for phase transitions or parallel action structure~\citep{huang2026toolomni,li2026deepagent}.
\textsc{ParaAgent} addresses both: \textsc{ToolEnv} provides realistic open-world API simulation, while hierarchical multi-level rewards supervise action format, phase transitions, and task outcomes.

\section{Methodology}
\subsection{Problem Statement}
\label{sec:problem}
We consider an agent operating in an open-world tool environment $\mathcal{L}$.
Given task $\tau$, the agent acts over $\mathcal{A} = \mathcal{A}^e \cup \mathcal{A}^o$:
\emph{exploration} phase $\mathcal{A}^e$ issues parallel retrieval queries to discover
relevant tools from $\mathcal{L}$, and \emph{execution} phase $\mathcal{A}^o$ invokes
the discovered tools concurrently under an inferred dependency.
The two phases form a structured
\emph{Exploration $\rightleftharpoons$ Execution} loop, producing a trajectory:
\begin{equation}
  \xi = (\mathcal{A}^e_1, o^e_1, \mathcal{A}^o_1, o^o_1, \ldots,
         \mathcal{A}^e_t, o^e_t, \mathcal{A}^o_t, o^o_t, \hat{y}),
  \label{eq:traj}
\end{equation}
where $o^e_t$ and $o^o_t$ are the retrieval and execution observations at loop
$t$, and $\hat{y}$ is the final answer.
The objective is to learn $\pi_\theta$ that maximizes
task success $\mathbb{E}_{\xi\sim\pi_\theta}
[\mathbf{1}[\hat{y}\text{ solves }\tau]]$.

\begin{figure*}[t]
    \centering
\includegraphics[width=0.93\linewidth]{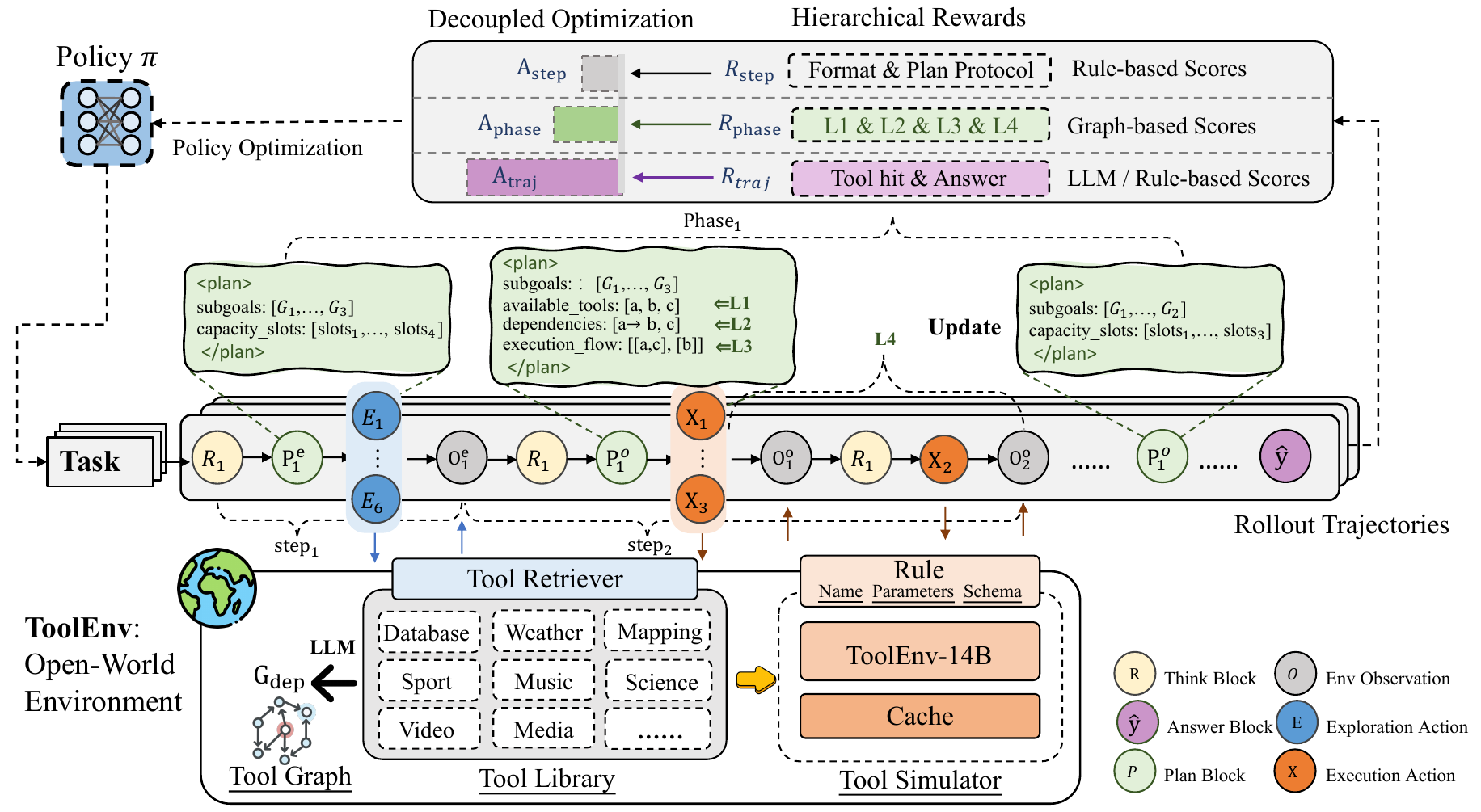}
    \caption{\textbf{Overview of \textsc{ParaAgent}.}
\textit{Middle}: The agent generates rollouts in \textsc{ToolEnv} through a structured parallel-action loop.
\textit{Top}: A hierarchical reward provides supervision at step, phase, and trajectory levels, optimizing policy $\pi$ via decoupled three advantages.
\textit{Bottom}: \textsc{ToolEnv} is an open-world environment comprising a real-world tool library, retriever, and three-layer simulator.}
    \label{fig:model}
\end{figure*}

\subsection{\textsc{ParaAct}: A Structured Parallel Action Loop Framework}
\label{sec:paraact}

\textsc{ParaAct} structures the interaction process by three layers of verbal reasoning (\textcolor{thinkcolor}{\texttt{<think>}}), structured plans (\textcolor{plancolor}{\texttt{<plan>}}), and parallel actions (\textcolor{toolcolor}{\texttt{<search\_tool>}} or \textcolor{toolcolor}{\texttt{<tool\_call>}}).
Exploration aims to identify capability gaps and determine which tools to search for, whereas execution aims to infer dependencies among acquired tools and schedule them efficiently.
This distinction yields different planning structures: exploration enumerates capability slots to be searched in parallel; execution schedules concurrent calls based on inferred dependencies. If tools are unusable or capability needs remain unmet, the agent starts a new loop, allowing the working tool set to expand incrementally. After multiple loops, the answer is given within the \textcolor{answercolor}{\texttt{<answer>}} and \textcolor{answercolor}{\texttt{</answer>}}.

\paragraph{Exploration Parallelism.}
$\mathcal{A}^e$ first reasons in \textcolor{thinkcolor}{\texttt{<think>}} and  \textcolor{thinkcolor}{\texttt{</think>}} tags to identify functional requirements the task needs, then organizes them into structured
\emph{capability slots} within the
\textcolor{plancolor}{\texttt{<plan>}}.
Next, the plan simultaneously drives parallel search actions $\{q_1, \ldots, q_j\}$, and each contains a description of the required functionality in a \textcolor{toolcolor}{\texttt{<search\_tool>}} and \textcolor{toolcolor}{\texttt{</search\_tool>}} tag.
Each $q_i$ retrieves the top-$k$ candidates from
$\mathcal{L}$, yielding up to $j \times k$ new tools at loop $t$ and updating
the accumulated working tool set as:
\begin{equation}
  \mathcal{B}_t = \mathcal{B}_{t-1} \cup \bigcup_{i=1}^{j}
  \operatorname{TopK}(\mathcal{L},\, q_i),
  \qquad \mathcal{B}_0 = \varnothing.
  \label{eq:explore}
\end{equation}
This enables targeted search over capability gaps rather than undirected
exploration. 
Based on the current slot coverage and accumulated observations,
the policy decides whether to continue exploration or transition.

\paragraph{Execution Parallelism.}
Some tools in $\mathcal{B}_t$ may be irrelevant or
unusable, \textit{e.g.}, due to a missing API key.
Therefore, $\mathcal{A}^o$ first reasons within
\textcolor{thinkcolor}{\texttt{<think>}} to identify usable
tools and infer their dependencies, and then organizes
them into a structured \textcolor{plancolor}{\texttt{<plan>}}
with three components:
(1)~\emph{Available tools}: the relevant subset of $\mathcal{B}_t$.
(2)~\emph{Dependencies}:
inter-call relationships modeled
as a DAG $G_t = (\mathcal{V}_t, \mathcal{D}_t)$, where
each node represents a uniquely identified call to an available tool,
and $(l_i, l_j) \in \mathcal{D}_t$ indicates that $l_j$ requires
an output or state change produced by $l_i$.
(3)~\emph{Execution flow}: a layer-wise topological schedule over $G_t$, where each layer $\mathcal{V}_s^{\parallel}$ contains all zero-in-degree tools in the residual graph and is dispatched concurrently via \textcolor{toolcolor}{\texttt{<tool\_call>}}.
\begin{equation}
  \mathcal{V}_s^{\parallel} = \{l \in \mathcal{V}_t :
  \deg^-(l) = 0\}, \quad
  G_t \leftarrow G_t \setminus \mathcal{V}_s^{\parallel},
  \label{eq:topo}
\end{equation}
where $\mathcal{V}_t$ denotes the call nodes remaining in $G_t$
and is updated as nodes are removed.
Each result $o^o$ feeds into dependent calls as input,
and execution continues until all planned calls are completed.
Structuring the plan explicitly elicits this reasoning from the model, while its phase-typed schema makes exploration and execution
distinguishable, enabling targeted optimization for each phase.

\subsection{\textsc{ParaAgent}: Learning the Parallel Action Loop}
\label{sec:learning}

To learn this structured parallel acting loop, we adopt a two-stage learning. Cold-start bootstrapping first instils the behavior from high-quality demonstrations. Online reinforcement learning, guided by multi-level advantage decoupling, then refines the policy through environment interaction.

\paragraph{Data Collection.}
\label{sec:graph}
\textbf{(1) Training data.}
We collect source training data from four datasets covering three tool types and diverse task structures: API-based tasks (ToolBench~\cite{qin2024toolllm}, Simia~\cite{li2025simulating}), MCP-based tasks
(Toucan~\cite{xu2025toucan}), and deep research tasks (AFM~\cite{li2025chain}).
The data comes from the training sets of these works.
We stratify tasks by difficulty, measured by the number of required tools, and bias the RL set toward harder, tool-intensive tasks.
\textbf{(2) Tool dependency graph.}
To support dependency-aware execution, we construct a global
\emph{tool dependency graph}
$\mathcal{G}_{\text{dep}}$ over the complete tool library
(\S\ref{sec:env}).
For each tool pair $(l_i, l_j)$ and each direction of their interaction, we adopt a voting scheme of four distinct LLM\footnote{kimi-k2.6,
qwen3-235b, and deepseek-v4-pro vote
independently, and gpt-5.1 breaks ties.} judges to label the relationship
between $l_j$  and $l_i$ as
\textit{HARD} (strongly dependent), \textit{SOFT}
(conditionally related), or \textit{NONE} (independent).
Graph statistics are summarized in Figure~\ref{fig:toolgraph-overview}(a).
See Appendix~\ref{app:data} for full data statistics and graph construction details.

\paragraph{Cold-Start Bootstrapping.}
\label{sec:coldstart}
We construct a \textit{state-sharing multi-agent system} to generate high-quality trajectories, providing a behavioral prior.
The system consists of five specialized agents coordinated by a shared state pool:
(1) \emph{Planner} decomposes tasks, generates slots or DAGs, and routes modes. (2)
\emph{Retriever} performs slot-guided parallel retrieval. (3) \emph{Executor}
traverses the DAG in topological order using Kahn's algorithm~\cite{kahn1962topological}
and dispatches independent tool calls in parallel. (4) \emph{Critic} provides transition supervision only during trajectory generation;
at inference time, the learned policy makes phase-transition decisions directly.
(5)  \emph{Answerer} synthesizes verified observations into the final answer.
All agents produce both reasoning and structured actions.
The \emph{Planner} routes the system across four modes: Exploration, Execution,
Refine, and Answer.
After each phase, the \emph{Critic} evaluates coverage to determine the next transition.
After format validation, the generated trajectories are
serialized and used to supervised fine-tune the base model.
See Appendix~\ref{app:coldstart} for full details of the multi-agent system design.

\begin{figure}[t]
\centering
\setlength{\parskip}{0pt}
\setlength{\abovecaptionskip}{4pt}
\begin{minipage}{0.88\linewidth}
\centering
\begin{minipage}[t]{0.43\linewidth}
\vspace{0pt}
\centering
\scalebox{0.94}{\begingroup
\footnotesize
\setlength{\tabcolsep}{3pt}
\renewcommand{\arraystretch}{0.94}
\begin{tabular}{@{}lr@{}}
\toprule
\textbf{$\mathcal{G}_{\text{dep}}$ statistic} & \textbf{Value} \\
\midrule
\#\,Tools (nodes)               & 50{,}011 \\
\#\,Tools w/ dependency         & 33{,}020 \\
\#\,Directed edges              & 147{,}503 \\
\quad--\,Strong (\textsc{hard}) & 43{,}087 \\
\quad--\,Soft (\textsc{soft})   & 104{,}416 \\
Mean out-degree (connected)     & 4.47 \\
Max out-degree                  & 232 \\
Max in-degree                   & 131 \\
\bottomrule
\end{tabular}%
\endgroup
}%
\end{minipage}\hfill%
\begin{minipage}[t]{0.54\linewidth}
\vspace{0pt}
\centering
\includegraphics[width=\linewidth]{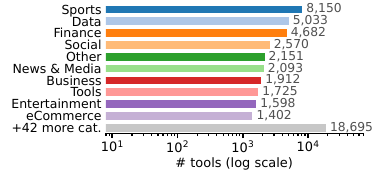}%
\end{minipage}%
\par\vspace{2pt}
\noindent
\begin{minipage}[t]{0.43\linewidth}
\centering
{\footnotesize \textbf{(a) Dependency graph }$\mathcal{G}_{\text{dep}}$}
\end{minipage}\hfill%
\begin{minipage}[t]{0.54\linewidth}
\centering
{\footnotesize \textbf{(b) Tool-library categories}}
\end{minipage}%
\end{minipage}
\caption{
(a) Statistics of $\mathcal{G}_{\text{dep}}$, with required \textsc{hard} dependencies and non-mandatory \textsc{soft} orderings.
(b) Top 10 of 52 categories by tool count in a long-tailed library.}
\label{fig:toolgraph-overview}
\end{figure}
\paragraph{Hierarchical Reward.}
Since our parallel action loop structures the
trajectory into phases and steps with distinct behavioral objectives, a single terminal
reward is insufficient: it cannot attribute failures to specific malformed action steps,
cannot guide the model to learn when to transition between exploration and execution
phases, and provides no signal for dependency-aware execution.
We therefore design rewards at three levels:

(1) \textbf{Step-level reward} $R_{\text{step}}$ evaluates structural and plan-protocol compliance of each step $s$, using two binary scores: $r_{\text{base}}^{(s)}$ checks basic structural tags
(\textit{e.g.}, \textcolor{thinkcolor}{\texttt{<think>}},
\textcolor{toolcolor}{\texttt{<tool\_call>}},
\textcolor{answercolor}{\texttt{<answer>}}).
$r_{\text{plan}}^{(s)}$ checks the required \textcolor{plancolor}{\texttt{<plan>}} block and its schema. The reward is averaged over all steps:
\begin{equation}
  R_{\text{step}} = \frac{1}{2|S|}\sum_{s \in S}
  \left(  r_{\text{base}}^{(s)} + r_{\text{plan}}^{(s)} \right),
  \label{eq:reward_step}
\end{equation}
where $r_{\text{plan}}^{(s)}=0$ for
the final \textcolor{answercolor}{\texttt{<answer>}} step.

(2) \textbf{Phase-level reward}
$R_{\text{phase}}$ scores
the structured parallel
action loop within each phase, identified by the
\textcolor{plancolor}{\texttt{<plan>}} block type.
Each phase is scored along four levels.
\textbf{L1} verifies that the model-declared available tools $\mathcal{V}_t$ are grounded in the retrieved tool set, and induces the subgraph $\mathcal{G}_{\text{sub}}$  of the global graph $\mathcal{G}_{\text{dep}}$ restricted to $\mathcal{V}_t$.
\textbf{L2} checks whether the dependency set
$\mathcal{D}$ matches the $\mathcal{G}_{\text{sub}}$.
\textbf{L3} verifies that the execution flow forms a valid topological order over
$\mathcal{G}_{\text{sub}}$.
\textbf{L4} measures whether actual
\textcolor{toolcolor}{\texttt{<tool\_call>}} follow the planned layer. $R_{\text{phase}}$ is averaged over all phases $\mathcal{P}$:
\begin{equation}
  R_{\text{phase}} = \frac{g_{\text{phase}}}{|\mathcal{P}|} \sum_{p \in \mathcal{P}}
  \mathbf{1}[\text{L1}]^{(p)}
  \Bigl(
    w_1
    + w_2 \cdot [\text{L2}]^{(p)}
    + w_3 \cdot [\text{L3}]^{(p)}
    + w_4 \cdot \text{L4}_{\text{ratio}}^{(p)}
  \Bigr).
  \label{eq:reward_phase}
\end{equation}
where $g_{\text{phase}}=\operatorname{clip}(R_{\text{ans}},0,1)$ gates self-generated plan rewards by the final outcome, $w_1,\ldots,w_4$ weight each level, and $\text{L4}_{\text{ratio}}^{(p)}$ denotes the fraction of ready scheduled tools successfully executed.

(3) \textbf{Trajectory-level reward}
$R_{\text{traj}}$  scores trajectory quality by combining tool hit against
ground-truth $\mathcal{T}^*$ and task  answer quality:
\begin{equation}
  R_{\text{traj}} =
  \sum_{m \in \{e,\,o\}}
  \gamma_m \cdot
  \frac{|\mathcal{T}_m \cap \mathcal{T}^*|}{|\mathcal{T}^*|}
  + \gamma_{\text{ans}} \cdot R_{\text{ans}},
  \label{eq:reward_traj}
\end{equation}
where $\mathcal{T}_e$ and $\mathcal{T}_o$ are retrieved and successfully invoked tools, respectively, and $\mathcal{T}^*$ is the reference set; $R_{\text{ans}}$ is evaluated by LLM- or rule-based methods.
See Appendix~\ref{app:reward} for full reward details.

\paragraph{Decoupled Optimization.}
For each task, we sample $N$ trajectories in \textsc{ToolEnv} and score them with the three reward components.
Summing these rewards before group-wise normalization may
obscure component-specific learning signals.
We therefore normalize each component independently
within the group~\cite{liu2026gdpo} and combine them into
a weighted advantage, with $\lambda_\tau > \lambda_p > \lambda_s$.
Let $\mu_k$ and $\sigma_k$ denote the group mean and standard
deviation of component $k \in \{\tau,p,s\}$.
We optimize the policy by minimizing:
\begin{equation}
  \mathcal{J} =
  -\mathbb{E}\!\left[
    \min\!\left(
      r_i \hat{A}_i,\;
      \operatorname{clip}\!\left(r_i,\, 1-\varepsilon,\, 1+\varepsilon\right) \hat{A}_i
    \right)
  \right]
  + \beta\, D_{\mathrm{KL}}(\pi_\theta \| \pi_{\text{ref}}),
  \label{eq:gdpo}
\end{equation}
where $\hat{A}_i = \sum_{k \in \{\tau, p, s\}} \lambda_k \frac{R_i^k - \mu_k}{\sigma_k}$ is the decoupled advantage, $r_i = \pi_\theta / \pi_{\theta_{\mathrm{old}}}$ is the sampling-policy importance ratio, $\varepsilon$ is the clipping threshold, and $\beta D_{\mathrm{KL}}$ penalizes divergence from the fixed reference policy $\pi_{\text{ref}}$.
This online learning scheme allows the agent to
continuously refine its exploration and execution strategies based on the environment
feedback.

\subsection{\textsc{ToolEnv}: Open-World Environment Construction}
\label{sec:env}

A realistic and controllable interaction environment is essential for training open-world tool agents, yet live APIs are impractical due to cost, instability, and privacy risks. Recent work shows that LLMs can effectively model tool behavior~\cite{guo2024stabletoolbench,li2025word,li2025simulating}. Following this, we build an LLM-driven tool environment that provides high-fidelity, controllable feedback for scalable training.
\paragraph{Data Collection.}
We collect and normalize 50{,}011 real-world tool interfaces spanning 52 vertical categories
(RapidAPI, OpenAPI Hub, MCP servers, \textit{etc.}) into a unified
OpenAPI-style library with four mandatory fields: tool name, description, parameter
schema, and category.
To train the tool simulator, we retain 241{,}949 real-world \textit{request--response} pairs from multiple sources after strict cleaning and alignment. (Appendix~\ref{app:data})
\paragraph{Tool Simulator.}
To ensure reliable and controllable simulation, the simulator consists of three
components.
(1) \textbf{Rule} first validates each input, checking tool
existence, required parameters, and schema constraints. It also injects controlled
errors, such as rate limits and network failures, to expose the agent to realistic
failure cases.
(2) \textbf{Response}, powered by LLM fine-tuned on collected
\textit{request--response} pairs, generates coherent and format-correct responses for
valid requests.
(3) \textbf{Cache} stores collected real-world responses
and reuses simulator outputs by writing them back to the cache, reducing inference
cost during training.
These three components balance simulation fidelity and training efficiency,
supporting stable and scalable agent training.

\section{Experimental Settings}
\label{sec:setting}
\begin{table*}[t]
\centering
\footnotesize
\setlength{\tabcolsep}{6.5pt}
\renewcommand{\arraystretch}{1.1}
\resizebox{\textwidth}{!}{%
\begin{tabular}{cl*{10}{c}}
\toprule
& & \multicolumn{8}{c}{\textbf{ToolBench}} & \multicolumn{2}{c}{\textbf{API-Bank}} \\
\cmidrule(lr){3-10}\cmidrule(lr){11-12}
& & \multicolumn{2}{c}{I1} & \multicolumn{2}{c}{I2} & \multicolumn{2}{c}{I3} & \multicolumn{2}{c}{\textbf{Avg}} & & \\
\cmidrule(lr){3-4}\cmidrule(lr){5-6}\cmidrule(lr){7-8}\cmidrule(lr){9-10}
\textbf{Action} & \textbf{Backbone} & Suc. & Path & Suc. & Path & Suc. & Path & Suc. & Path & Suc. & Acc. \\
\midrule
\multicolumn{12}{c}{\textit{Exploration-then-Execution}} \\[1pt]
ReAct    & GPT-4.1             & 43.2 & 28.9{\scriptsize$\pm$0.4} & 29.6 & 17.3{\scriptsize$\pm$1.2} & 29.5 & 10.2{\scriptsize$\pm$2.4} & 34.1 & 18.8{\scriptsize$\pm$0.3} & \cellcolor[HTML]{e4f8e4} 20.0 & 44.3{\scriptsize$\pm$1.1} \\
\cdashline{1-12}[2pt/2pt]
\multirow{2}{1.8cm}{\centering DFS}
 & GPT-4.1                     & 45.1 & 29.2{\scriptsize$\pm$0.3} & 28.7 & 17.4{\scriptsize$\pm$1.3} & 27.9 & 9.6{\scriptsize$\pm$2.1} & 33.9 & 18.7{\scriptsize$\pm$0.2} & \cellcolor[HTML]{e4f8e4} 20.0 & 45.4{\scriptsize$\pm$0.5} \\
 & T-LLaMA$_{7B}$                  & 15.6 & 32.7{\scriptsize$\pm$0.4} & 6.5 & 21.0{\scriptsize$\pm$0.1} & 11.5 & 12.5{\scriptsize$\pm$0.6} & 11.2 & 22.1{\scriptsize$\pm$0.3} & 2.0 & 21.4{\scriptsize$\pm$0.0} \\
\cdashline{1-12}[2pt/2pt]
\multirow{2}{1.8cm}{\centering Parallel}
 & GPT-4.1                     & 37.1 & 28.0{\scriptsize$\pm$0.1} & 28.7 & 20.9{\scriptsize$\pm$1.7} & 18.0 & 10.7{\scriptsize$\pm$0.2} & 28.0 & 19.8{\scriptsize$\pm$0.6} & \cellcolor[HTML]{a5daa6} 22.0 & \cellcolor[HTML]{a5daa6} 58.4{\scriptsize$\pm$2.7} \\
 & DTA$_{8B}$                      & 16.2 & 32.4{\scriptsize$\pm$0.4} & 8.3 & 20.4{\scriptsize$\pm$0.3} & 16.4 & 15.3{\scriptsize$\pm$0.4} & 13.6 & 22.7{\scriptsize$\pm$0.4} & 2.0 & 17.6{\scriptsize$\pm$0.0} \\
\cdashline{1-12}[2pt/2pt]
ReWOO    & GPT-4.1             & 44.1 & 36.6{\scriptsize$\pm$0.7} & 22.6 & 22.7{\scriptsize$\pm$0.3} & \cellcolor[HTML]{e4f8e4} 44.3 & 14.0{\scriptsize$\pm$0.4} & 37.0 & 24.4{\scriptsize$\pm$0.3} & 12.0 & 29.0{\scriptsize$\pm$0.0} \\
\cdashline{1-12}[2pt/2pt]
ToolOmni & ToolOmni$_{4B}$            & 29.1 & \cellcolor[HTML]{a5daa6} 47.2{\scriptsize$\pm$0.1} & 28.3 & \cellcolor[HTML]{a5daa6} 39.4{\scriptsize$\pm$0.1} & 21.3 & \cellcolor[HTML]{a5daa6} 30.7{\scriptsize$\pm$0.3} & 26.2 & \cellcolor[HTML]{a5daa6} 39.1{\scriptsize$\pm$0.1} & \cellcolor[HTML]{e4f8e4} 20.0 & 46.6{\scriptsize$\pm$0.0} \\
\midrule
\multicolumn{12}{c}{\textit{Exploration-and-Execution}} \\[1pt]
\multirow{3}{1.8cm}{\centering ReAct}
 & GPT-4.1                     & \cellcolor[HTML]{e4f8e4} 62.2 & 34.7{\scriptsize$\pm$0.6} & 55.7 & 28.2{\scriptsize$\pm$1.6} & 41.0 & 14.3{\scriptsize$\pm$0.3} & \cellcolor[HTML]{e4f8e4} 53.0 & 25.7{\scriptsize$\pm$0.4} & \cellcolor[HTML]{a5daa6} 22.0 & 57.6{\scriptsize$\pm$1.6} \\
 & Qwen3$_{32B}$           & 46.8 & 29.1{\scriptsize$\pm$0.4} & 48.3 & 27.7{\scriptsize$\pm$0.9} & 27.9 & 18.8{\scriptsize$\pm$1.8} & 41.0 & 25.2{\scriptsize$\pm$0.5} & \cellcolor[HTML]{a5daa6} 22.0 & 53.4{\scriptsize$\pm$1.1} \\
 & Qwen3$_{4B}$            & 33.8 & 30.8{\scriptsize$\pm$0.3} & 36.1 & 25.9{\scriptsize$\pm$0.0} & 23.0 & 8.7{\scriptsize$\pm$0.9} & 30.9 & 21.8{\scriptsize$\pm$0.2} & 14.0 & 46.2{\scriptsize$\pm$5.9} \\
\cdashline{1-12}[2pt/2pt]
\multirow{3}{1.8cm}{\centering DFS}
 & GPT-4.1                     & \cellcolor[HTML]{a5daa6} 62.7 & 33.4{\scriptsize$\pm$0.2} & \cellcolor[HTML]{e4f8e4} 56.5 & 27.7{\scriptsize$\pm$0.8} & 37.7 & 13.3{\scriptsize$\pm$1.5} & 52.3 & 24.8{\scriptsize$\pm$1.7} & \cellcolor[HTML]{a5daa6} 22.0 & 53.4{\scriptsize$\pm$3.2} \\
 & Qwen3$_{32B}$           & 50.6 & 29.9{\scriptsize$\pm$0.1} & 47.4 & 29.1{\scriptsize$\pm$0.1} & 39.3 & 21.4{\scriptsize$\pm$2.6} & 45.8 & 26.8{\scriptsize$\pm$1.0} & \cellcolor[HTML]{a5daa6} 22.0 & 56.5{\scriptsize$\pm$1.1} \\
 & Qwen3$_{4B}$            & 47.5 & 32.4{\scriptsize$\pm$0.8} & 53.0 & 28.1{\scriptsize$\pm$1.6} & 42.6 & 10.5{\scriptsize$\pm$2.0} & 47.7 & 23.7{\scriptsize$\pm$0.9} & 18.0 & 49.6{\scriptsize$\pm$3.2} \\
\cdashline{1-12}[2pt/2pt]
\multirow{3}{1.8cm}{\centering Parallel}
 & GPT-4.1                     & \cellcolor[HTML]{5eb95f} 62.9 & 34.2{\scriptsize$\pm$1.3} & \cellcolor[HTML]{a5daa6} 57.8 & 31.0{\scriptsize$\pm$0.7} & \cellcolor[HTML]{a5daa6} 45.9 & 21.4{\scriptsize$\pm$0.7} & \cellcolor[HTML]{a5daa6} 55.5 & 28.9{\scriptsize$\pm$0.5} & \cellcolor[HTML]{a5daa6} 22.0 & 56.5{\scriptsize$\pm$1.3} \\
 & Qwen3$_{32B}$           & 35.7 & 23.7{\scriptsize$\pm$0.6} & 33.0 & 19.4{\scriptsize$\pm$0.6} & 26.2 & 11.3{\scriptsize$\pm$1.2} & 31.6 & 18.1{\scriptsize$\pm$0.0} & 10.0 & 48.1{\scriptsize$\pm$1.3} \\
 & Qwen3$_{4B}$            & 31.0 & 26.2{\scriptsize$\pm$0.9} & 32.2 & 21.9{\scriptsize$\pm$1.2} & 24.6 & 7.2{\scriptsize$\pm$0.2} & 29.3 & 18.5{\scriptsize$\pm$0.1} & 18.0 & 28.2{\scriptsize$\pm$0.8} \\
\cdashline{1-12}[2pt/2pt]
DeepAgent & DeepAgent$_{32B}$            & 39.7 & \cellcolor[HTML]{e4f8e4} 39.4{\scriptsize$\pm$0.3} & 37.8 & \cellcolor[HTML]{e4f8e4} 34.2{\scriptsize$\pm$1.2} & 27.9 & \cellcolor[HTML]{e4f8e4} 22.0{\scriptsize$\pm$0.8} & 35.1 & \cellcolor[HTML]{e4f8e4} 31.9{\scriptsize$\pm$0.4} & \cellcolor[HTML]{5eb95f} 24.0 & \cellcolor[HTML]{5eb95f} 59.9{\scriptsize$\pm$2.7} \\
\midrule
\multicolumn{12}{c}{\textit{Ours}} \\[1pt]
\textsc{ParaAct} & \textsc{ParaAgent}$_{4B}$    & 58.4 & \cellcolor[HTML]{5eb95f} 47.8{\scriptsize$\pm$0.8} & \cellcolor[HTML]{5eb95f} 65.7 & \cellcolor[HTML]{5eb95f} 46.5{\scriptsize$\pm$0.2} & \cellcolor[HTML]{5eb95f} 55.7 & \cellcolor[HTML]{5eb95f} 34.5{\scriptsize$\pm$1.0} & \cellcolor[HTML]{5eb95f} 59.9 & \cellcolor[HTML]{5eb95f} 42.9{\scriptsize$\pm$0.5} & \cellcolor[HTML]{5eb95f} 24.0 & \cellcolor[HTML]{e4f8e4} 58.0{\scriptsize$\pm$0.0} \\
\bottomrule
\end{tabular}
}
\caption{\textbf{Main results across paradigm $\times$ action strategy.}
Suc.: Pass@2 (\%). Path: mean$\pm$sample SD (\%).
Cell shading marks the \colorbox[HTML]{5eb95f}{best}, \colorbox[HTML]{a5daa6}{second}, and \colorbox[HTML]{e4f8e4}{third} per column.}
\label{tab:main}
\end{table*}

\paragraph{\textbf{Tasks and Datasets.}}
We evaluate on two open-world tool benchmarks.
\textbf{ToolBench}~\cite{guo2024stabletoolbench}  contains 765 tasks over 3{,}583 APIs, with increasingly complex splits: \textbf{I1} (474 single-tool tasks), \textbf{I2} (230 within-category multi-tool tasks), and \textbf{I3} (61 intra-collection multi-tool tasks).
\textbf{API-Bank} evaluates joint API planning, retrieval, and calling~\cite{li2023api}; we use its Level-3 split of 50 tasks over 73 APIs.
\emph{Success} (Suc.) measures task completion; ToolBench \emph{Tool-Path} (Path) measures ground-truth tool hit coverage, while API-Bank \emph{API Accuracy} (Acc.) uses the official scorer.

\paragraph{\textbf{Baselines.}}
We compare two \textit{\textbf{exploration-execution paradigms}} across representative \textit{\textbf{action strategies}}.
\textbf{Explo\-ra\-tion-then-Execution} includes ReAct~\cite{yao2023react}, DFS~\cite{qin2024toolllm}, Parallel~\cite{zhu2025divide}, ReWOO~\citep{xu2023rewoo}, and ToolOmni~\citep{huang2026toolomni}, with specialized models T-LLaMA-7B, DTA-8B.
\textbf{Exploration-and-Execution} includes ReAct~\cite{yao2023react}, DFS~\cite{qin2024toolllm}, and Parallel~\cite{zhu2025divide} with GPT-4.1\footnote{\texttt{gpt-4.1-2025-04-14}}~\cite{achiam2023gpt}, Qwen3-32B, and Qwen3-4B~\cite{yang2025qwen3}, plus DeepAgent$_{32B}$~\citep{li2026deepagent}.

\paragraph{\textbf{Implementation Details.}}
\textsc{ParaAgent} uses Qwen3-4B~\cite{yang2025qwen3} (SFT: 3 epochs, lr $1.0\times10^{-5}$; RL: 2 epochs over 16{,}384 prompts, $N{=}8$, $\lambda_\tau{=}1.0$, $\lambda_p{=}0.5$, $\lambda_s{=}0.25$).
\textsc{ToolEnv} uses Qwen3-14B~\cite{yang2025qwen3} (SFT: 2 epochs, lr $1.0\times10^{-5}$).
We use Qwen3-235B-A22B-Instruct-2507 as the RL judge and bge-large-en-v1.5~\cite{zhang2023retrieve} as the retriever with top-3 candidates.
All experiments run on 16 $\times$ NVIDIA H200-141GB GPUs.

Full details on datasets, baselines, and implementation are provided in Appendix~\ref{app:setup}.

\section{Experimental Results}
\label{sec:main}

\begin{figure}[t]
\centering
\includegraphics[width=\linewidth]{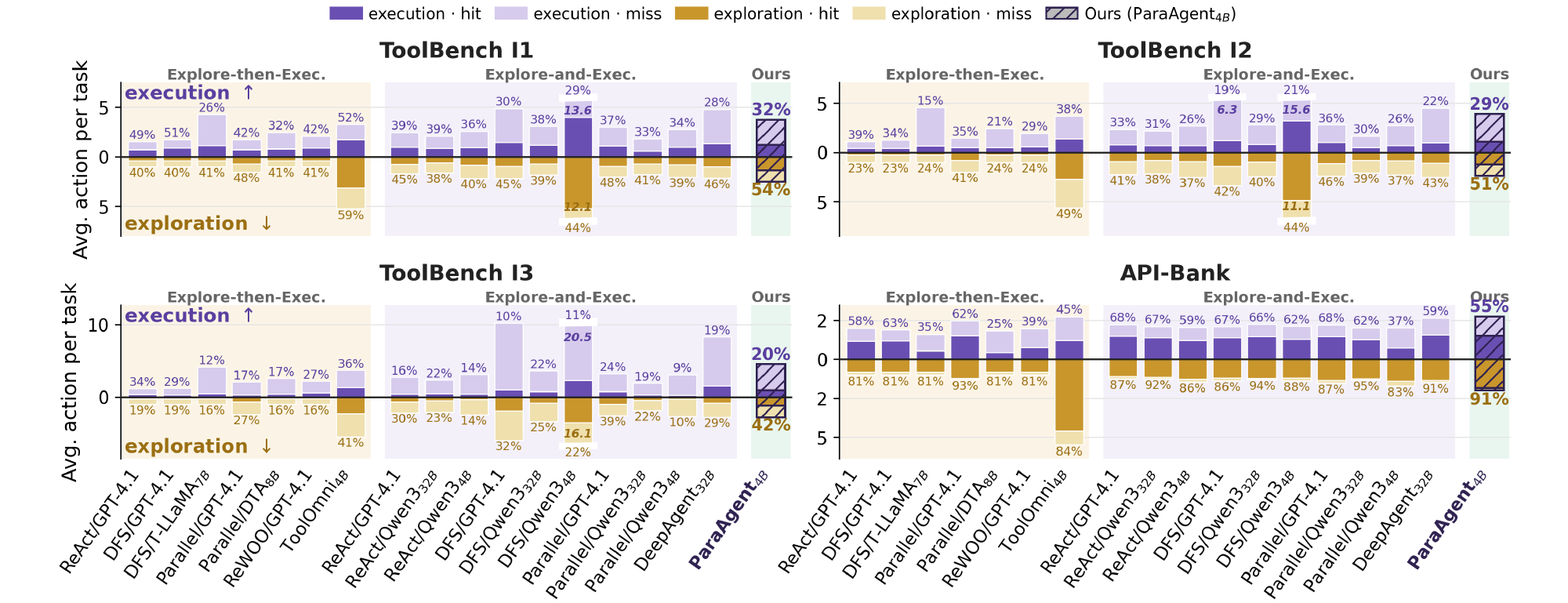}
\caption{\textbf{Action behavior per task.} Above zero: calls (\%: hits/calls; API-Bank uses exact match). Below zero: searches (\%: gold-API coverage). Two-run means.}
\label{fig:toolcall}
\end{figure}

The experiments answer five questions. For
\textbf{Performance--efficiency tradeoff:} RQ1 compares \textsc{ParaAgent} with existing agents (Table~\ref{tab:main}) and RQ2 measures its cost in turns and tokens (Figure~\ref{fig:tradeoff}).
 For \textbf{Structure:} RQ3 analyzes how actions are allocated (Figure~\ref{fig:toolcall}).
For \textbf{Learning:} RQ4 and RQ5 ablate the training and the reward components (Table~\ref{tab:ablation}).

\subsubsection*{\rqtag{RQ1}~How does \textsc{ParaAgent} compare with state-of-the-art baselines?}

Table~\ref{tab:main} compares \textsc{ParaAgent} with 17 baselines covering both paradigms, diverse action strategies, and backbones from 4B to GPT-4.1.
\textsc{ParaAgent}$_{4B}$ achieves the best average ToolBench success and great Tool-Path.
\textbf{\emph{(1) Across paradigms}}, Exploration-then-Execution trails Exploration-and-Execution by $18$--$28$ points under the same GPT-4.1 backbone and action strategy.
Once the tool set is fixed, retrieval errors cannot be corrected during execution: even ToolOmni, which has the highest baseline Path score, completes only $26.2\%$ of tasks.
\textbf{\emph{(2) Across action strategies}} at the same 4B scale, parallelism alone does not help: dependency-blind Parallel is no better than serial ReAct.
Tree search (DFS) is the strongest 4B baseline because it can backtrack, but it takes the most turns of all systems.
\textsc{ParaAgent} exceeds it by $12.2$ points with a dependency-aware DAG that parallelizes only independent calls.
\textbf{\emph{(3) Across scales}}, all larger baselines fall below \textsc{ParaAgent}$_{4B}$ on average, including T-LLaMA$_{7B}$, DTA$_{8B}$, DeepAgent$_{32B}$, and every Qwen3-32B and GPT-4.1 configuration.
Scaling the Qwen3 backbone from 4B to 32B changes success by $-1.9$ to $+10.1$ points, less than the $12.2$-point gain of \textsc{ParaAgent} over the best 4B baseline.
The advantage grows with task complexity: on single-tool I1, where there is less to coordinate, the GPT-4.1 Exploration-and-Execution baselines lead; on multi-tool I2 and I3, \textsc{ParaAgent} leads them.
Overall, the gains come from how actions are structured rather than from model scale, and they are largest on multi-tool tasks.

\noindent\begin{minipage}{\textwidth}
\subsection*{\rqtag{RQ2}~Does the higher success cost more turns or tokens?}

\noindent\begin{minipage}[t]{0.47\textwidth}
\vspace{0pt}
Figure~\ref{fig:tradeoff} plots task completion against two cost axes, interaction turns and output tokens per task.
\textbf{\emph{(1) Exploration-then-Execution is cheap but weak}}: most agents finish within five turns but reach at most $24.0\%$.
\textbf{\emph{(2) Exploration-and-Execution spends more but plateaus}}: DFS/Qwen3-4B takes $19.5$ turns for $25.7\%$, and DeepAgent$_{32B}$ emits $4.2$k tokens for $22.7\%$.
\textbf{\emph{(3) Structured parallelism}} plans dependencies before batching calls, unlike native parallelism (Parallel/GPT-4.1), and packs about two actions into each turn (at most $1.1$ for baselines), so \textsc{ParaAgent}$_{4B}$ reaches the highest success in only $3.74$ turns with affordable cost.

\end{minipage}\hfill
\begin{minipage}[t]{0.5\textwidth}
\vspace{0pt}
\centering
\includegraphics[width=0.95\linewidth]{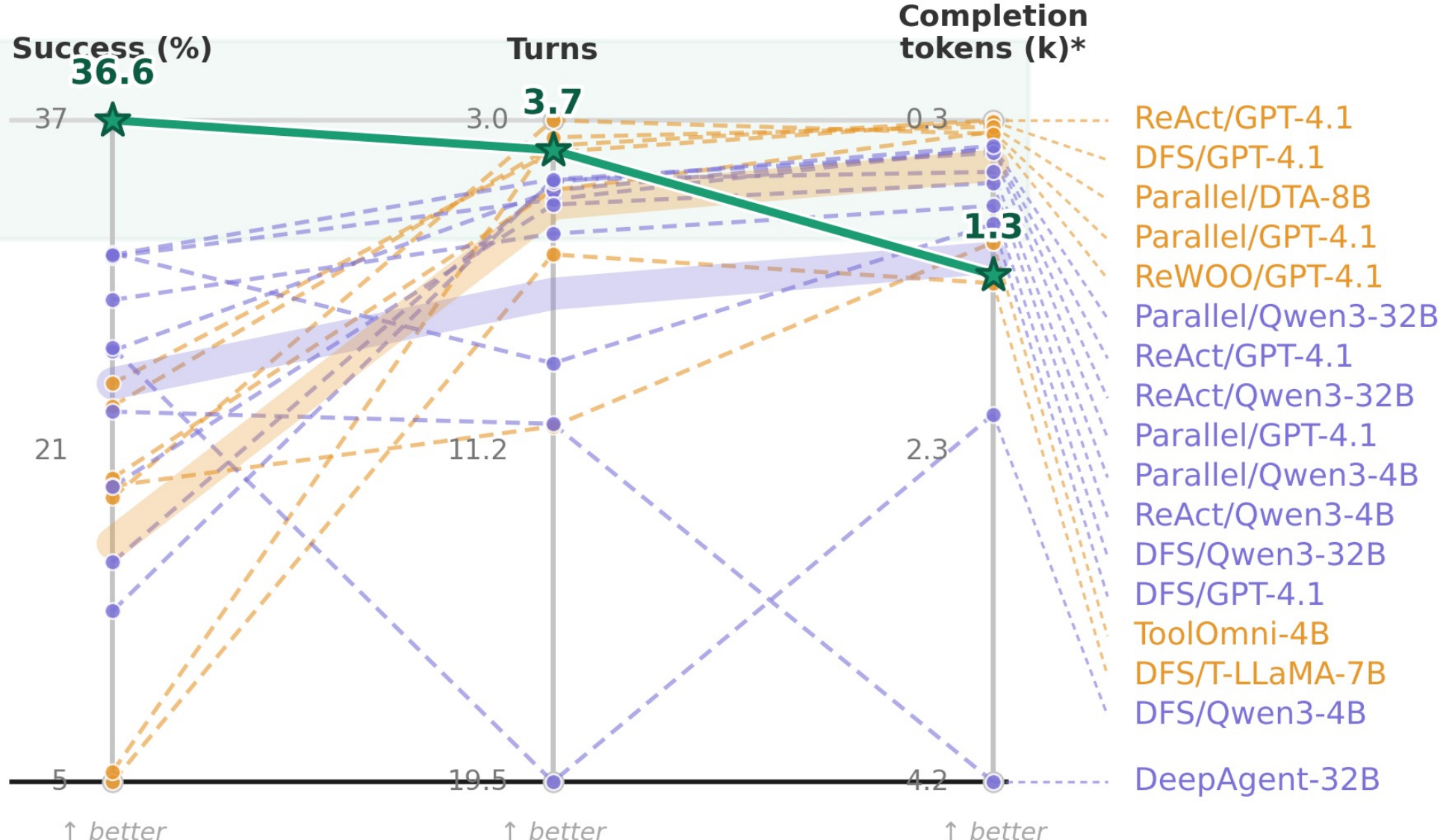}
  \captionof{figure}{\textbf{Performance-efficiency trade-off} averaged over two runs and both benchmarks.}
  \label{fig:tradeoff}
\end{minipage}
\end{minipage}
\par\medskip

\subsection*{\rqtag{RQ3}~Do the gains come from better action decisions or from more actions?}

Figure~\ref{fig:toolcall} separates exploration (\texttt{search\_tool})  from execution (\texttt{tool\_call}), measuring retrieval coverage as the fraction of gold tools surfaced by search and hit rate as the fraction of calls invoking a gold tool.
\textsc{ParaAgent}$_{4B}$ uses only $6.6$ actions per ToolBench task, fewer than ToolOmni and less than one quarter of same-backbone DFS, yet achieves the highest success.
Its exploration is efficient rather than exhaustive, matching ToolOmni's coverage with less than half the searches and leading on I2 and I3. Its execution is recoverable rather than maximally precise: despite a moderate hit rate, interleaved search and execution allow it to recover missing tools on the fly.
On API-Bank, \textsc{ParaAgent}$_{4B}$ ties for the best success, showing that completion depends on how calls are organized and called.
The gains come from high-quality action decisions, efficient exploration paired with recoverable execution, rather than from acting more.

\begin{table}[t]
\centering
\begingroup
\scriptsize
\setlength{\tabcolsep}{0pt}
\renewcommand{\arraystretch}{1.1}
\begin{minipage}[t]{0.42\linewidth}
\vspace{0pt}
\centering
\begin{tabular}{@{}p{0.36\linewidth}*{4}{>{\centering\arraybackslash}p{0.16\linewidth}}@{}}
\toprule
& \multicolumn{2}{c}{\textbf{ToolBench}} & \multicolumn{2}{c}{\textbf{API-Bank}} \\
\cmidrule(lr){2-3}\cmidrule(lr){4-5}
\textbf{Method} & Suc. & Path & Suc. & Acc. \\
\midrule
Backbone        & 22.13 & 23.49 & 16.00 & 35.88 \\
+ SFT            & 31.90 & 37.54 & 20.00 & 43.51 \\
\addlinespace[2pt]
\cdashline{1-5}[2pt/2pt]
\addlinespace[2pt]
+  SFT \& RL(GRPO)                  & 47.02 & \textbf{46.51} & 14.00 & 52.67 \\
\textbf{+  SFT \& RL (Ours)}        & \textbf{49.66} & 42.54 & \textbf{24.00} & \textbf{58.02} \\
\bottomrule
\end{tabular}
\par\vspace{2pt}
\textbf{(a) Learning strategies}
\end{minipage}\hspace{0.03\linewidth}%
\begin{minipage}[t]{0.42\linewidth}
\vspace{0pt}
\centering
\begin{tabular}{@{}p{0.25\linewidth}*{2}{>{\centering\arraybackslash}p{0.235\linewidth}}*{2}{>{\centering\arraybackslash}p{0.14\linewidth}}@{}}
\toprule
& \multicolumn{2}{c}{\textbf{Success ($\uparrow$)}} & \multicolumn{2}{c}{\textbf{Efficiency ($\downarrow$)}} \\
\cmidrule(lr){2-3}\cmidrule(lr){4-5}
\textbf{Method} & ToolBench & API-Bank & Turns & Tok. \\
\midrule
\textbf{Full (Ours)}        & \textbf{49.66} & \textbf{24.00} & 3.76 & 1.26 \\
\addlinespace[2pt]
\cdashline{1-5}[2pt/2pt]
\addlinespace[2pt]
w/o $g_{\text{phase}}$ & 40.56 & 20.00 & \textbf{3.43} & \textbf{1.14} \\
w/o $R_{\text{step}}$        & 44.21 & 14.00 & 3.84 & 1.29 \\
w/o $R_{\text{phase}}$       & 44.56 & 18.00 & 3.52 & 1.20 \\
\bottomrule
\end{tabular}
\par\vspace{2pt}
\textbf{(b) Hierarchical reward}
\end{minipage}
\par
\endgroup
\caption{\textbf{Ablation analysis of learning strategies and hierarchical reward}.
Suc.: Pass@1 (\%); Acc.: API accuracy (\%); Turns and Tok.\ (output, k): averaged over both benchmarks.}
\label{tab:ablation}
\end{table}

\subsection*{\rqtag{RQ4}~How do the training stages and decoupled normalization contribute?}

Table~\ref{tab:ablation} (a) progressively adds each training stage to Qwen3-4B and compares multi-level advantage decoupling with GRPO~\cite{shao2024deepseekmath} that aggregates the same hierarchical rewards before normalization. Cold-start SFT improves ToolBench success by $9.77$ points and API accuracy by $7.63$ points, showing that demonstrations initialize the loop. RL yields larger gains: decoupled optimization further improves ToolBench success by $17.76$ points and API accuracy by $14.51$ points. With identical rewards, the gap between the two RL variants isolates the effect of decoupling: it improves API-Bank success by $10.00$ points and API accuracy by $5.35$ points, while aggregated GRPO even underperforms SFT in success. Its higher Path score but lower success further suggests that reward aggregation obscures learning signals and favors tool coverage over task completion. Overall, demonstrations initialize the loop, interaction refines structural decisions, and decoupled optimization better aligns action organization with task outcomes.

\subsection*{\rqtag{RQ5}~Does each reward component contribute to success?}
Table~\ref{tab:ablation} (b) reports ablations of the hierarchical reward components from the Full model, including the step reward $R_{\text{step}}$ for plan-protocol and format compliance, the phase reward $R_{\text{phase}}$ for evaluating planning and execution against the dependency subgraph of declared tools, and the outcome gate $g_{\text{phase}}$ on positive phase rewards.
Removing any component reduces success on both benchmarks, supporting our design that both action organization and final outcomes should be rewarded.
Removing $R_{\text{phase}}$ decreases performance by $5.10$ and $6.00$ points while reducing interaction to $3.52$ turns: without graph supervision, the policy may compress dependency chains and issue dependent calls before their inputs are ready, saving turns at the cost of success.
Removing the gate causes the largest drop on ToolBench, as self-generated plans may be internally consistent yet omit task-required tools, while the gate prevents such plans from receiving rewards.
Overall, all levels of supervision contribute, demonstrating the effectiveness of our multi-level reward design.

\section{Conclusion}
We studied how LLM agents should coordinate exploration and execution in open-world tool environments, where existing paradigms face a performance--efficiency tradeoff.
We introduced \textsc{ParaAct}, a structured parallel-action loop that couples phase-level adaptation with action-level parallelism, and \textsc{ParaAgent}, which learns this loop through multi-agent cold-start supervision and reinforcement learning with decoupled hierarchical rewards in \textsc{ToolEnv}.
On ToolBench and API-Bank, \textsc{ParaAgent}$_{4B}$ achieves the best average success, outperforming all baselines including GPT-4.1-based systems, with the largest gains on multi-tool tasks and fewer than four interaction turns per task.
Our analyses indicate that these gains stem from better action decisions rather than more interaction, suggesting that how an agent organizes its actions can matter as much as backbone scale.
Limitations: we train only a 4B policy, rely on a simulated tool environment and LLM judges for training signals.
Future work includes scaling to larger backbones and extending the loop to Skill and Harness ecosystems.

\bibliographystyle{assets/plainnat}
\bibliography{references}

\begin{thebibliography}{51}
\providecommand{\natexlab}[1]{#1}
\providecommand{\url}[1]{\texttt{#1}}
\expandafter\ifx\csname urlstyle\endcsname\relax
  \providecommand{\doi}[1]{doi: #1}\else
  \providecommand{\doi}{doi: \begingroup \urlstyle{rm}\Url}\fi

\bibitem[Achiam et~al.(2023)Achiam, Adler, Agarwal, Ahmad, Akkaya, Aleman, Almeida, Altenschmidt, Altman, Anadkat, et~al.]{achiam2023gpt}
Josh Achiam, Steven Adler, Sandhini Agarwal, Lama Ahmad, Ilge Akkaya, Florencia~Leoni Aleman, Diogo Almeida, Janko Altenschmidt, Sam Altman, Shyamal Anadkat, et~al.
\newblock Gpt-4 technical report.
\newblock \emph{arXiv preprint arXiv:2303.08774}, 2023.

\bibitem[Chen et~al.(2024)Chen, Liu, Wang, Zhang, Liu, Lin, Chen, and Zhao]{chen2024agent}
Zehui Chen, Kuikun Liu, Qiuchen Wang, Wenwei Zhang, Jiangning Liu, Dahua Lin, Kai Chen, and Feng Zhao.
\newblock Agent-flan: Designing data and methods of effective agent tuning for large language models.
\newblock In \emph{Findings of the Association for Computational Linguistics: ACL 2024}, pages 9354--9366, 2024.

\bibitem[Cheng et~al.(2026)Cheng, Li, Yue, Lu, Qiu, Tian, Zhang, Qian, Wang, Chen, et~al.]{cheng2026theory}
Zihao Cheng, Yixia Li, Shengbin Yue, Yuheng Lu, Yuli Qiu, Yanzhi Tian, Haohuinan Zhang, Cheng Qian, Hongru Wang, Guanhua Chen, et~al.
\newblock Theory of agent: The science of internalization and externalization for llm-based agents.
\newblock 2026.

\bibitem[Feng et~al.(2025)Feng, Huang, Qu, Zhang, Qin, Zhong, Jiang, Chi, and Zhong]{feng2025retool}
Jiazhan Feng, Shijue Huang, Xingwei Qu, Ge~Zhang, Yujia Qin, Baoquan Zhong, Chengquan Jiang, Jinxin Chi, and Wanjun Zhong.
\newblock Retool: Reinforcement learning for strategic tool use in llms.
\newblock \emph{arXiv preprint arXiv:2504.11536}, 2025.

\bibitem[Guo et~al.(2025)Guo, Yang, Zhang, Song, Wang, Zhu, Xu, Zhang, Ma, Bi, et~al.]{guo2025deepseek}
Daya Guo, Dejian Yang, Haowei Zhang, Junxiao Song, Peiyi Wang, Qihao Zhu, Runxin Xu, Ruoyu Zhang, Shirong Ma, Xiao Bi, et~al.
\newblock Deepseek-r1: Incentivizing reasoning capability in llms via reinforcement learning.
\newblock \emph{arXiv preprint arXiv:2501.12948}, 2025.

\bibitem[Guo et~al.(2024)Guo, Cheng, Wang, Liang, Qin, Li, Liu, Sun, and Liu]{guo2024stabletoolbench}
Zhicheng Guo, Sijie Cheng, Hao Wang, Shihao Liang, Yujia Qin, Peng Li, Zhiyuan Liu, Maosong Sun, and Yang Liu.
\newblock Stabletoolbench: Towards stable large-scale benchmarking on tool learning of large language models.
\newblock In \emph{Findings of the Association for Computational Linguistics: ACL 2024}, pages 11143--11156, 2024.

\bibitem[Hou et~al.(2025)Hou, Zhao, Wang, and Wang]{hou2025model}
Xinyi Hou, Yanjie Zhao, Shenao Wang, and Haoyu Wang.
\newblock Model context protocol (mcp): Landscape, security threats, and future research directions.
\newblock \emph{ACM Transactions on Software Engineering and Methodology}, 2025.

\bibitem[Huang et~al.(2026)Huang, Zhang, Hu, and Zhang]{huang2026toolomni}
Shouzheng Huang, Meishan Zhang, Baotian Hu, and Min Zhang.
\newblock Toolomni: Enabling open-world tool use via agentic learning with proactive retrieval and grounded execution.
\newblock \emph{arXiv preprint arXiv:2604.13787}, 2026.

\bibitem[Jia et~al.(2026)Jia, Yue, Chen, Wang, Liu, Li, Song, and Wei]{jia2026ready}
Zheng Jia, Shengbin Yue, Wei Chen, Siyuan Wang, Yidong Liu, Zejun Li, Yun Song, and Zhongyu Wei.
\newblock Ready jurist one: Benchmarking language agents for legal intelligence in dynamic environments.
\newblock In \emph{Proceedings of the 64th Annual Meeting of the Association for Computational Linguistics (Volume 1: Long Papers)}, pages 10351--10376, 2026.

\bibitem[Jin et~al.(2025)Jin, Zeng, Yue, Yoon, Arik, Wang, Zamani, and Han]{jin2025search}
Bowen Jin, Hansi Zeng, Zhenrui Yue, Jinsung Yoon, Sercan Arik, Dong Wang, Hamed Zamani, and Jiawei Han.
\newblock Search-r1: Training llms to reason and leverage search engines with reinforcement learning.
\newblock \emph{arXiv preprint arXiv:2503.09516}, 2025.

\bibitem[Kachuee et~al.(2025)Kachuee, Ahuja, Kumar, Xu, and Liu]{kachuee2025improving}
Mohammad Kachuee, Sarthak Ahuja, Vaibhav Kumar, Puyang Xu, and Xiaohu Liu.
\newblock Improving tool retrieval by leveraging large language models for query generation.
\newblock In \emph{Proceedings of the 31st International Conference on Computational Linguistics: Industry Track}, pages 29--38, 2025.

\bibitem[Kahn(1962)]{kahn1962topological}
Arthur~B Kahn.
\newblock Topological sorting of large networks.
\newblock \emph{Communications of the ACM}, 5\penalty0 (11):\penalty0 558--562, 1962.

\bibitem[Kwon et~al.(2023)Kwon, Li, Zhuang, Sheng, Zheng, Yu, Gonzalez, Zhang, and Stoica]{kwon2023efficient}
Woosuk Kwon, Zhuohan Li, Siyuan Zhuang, Ying Sheng, Lianmin Zheng, Cody~Hao Yu, Joseph Gonzalez, Hao Zhang, and Ion Stoica.
\newblock Efficient memory management for large language model serving with pagedattention.
\newblock In \emph{Proceedings of the 29th Symposium on Operating Systems Principles}, pages 611--626, 2023.

\bibitem[Li et~al.(2025{\natexlab{a}})Li, Zhang, Yin, Zhang, Ou, Wu, Yin, Li, Tao, Wang, et~al.]{li2025websailor}
Kuan Li, Zhongwang Zhang, Huifeng Yin, Liwen Zhang, Litu Ou, Jialong Wu, Wenbiao Yin, Baixuan Li, Zhengwei Tao, Xinyu Wang, et~al.
\newblock Websailor: Navigating super-human reasoning for web agent.
\newblock \emph{arXiv preprint arXiv:2507.02592}, 2025{\natexlab{a}}.

\bibitem[Li et~al.(2023)Li, Zhao, Yu, Song, Li, Yu, Li, Huang, and Li]{li2023api}
Minghao Li, Yingxiu Zhao, Bowen Yu, Feifan Song, Hangyu Li, Haiyang Yu, Zhoujun Li, Fei Huang, and Yongbin Li.
\newblock Api-bank: A comprehensive benchmark for tool-augmented llms.
\newblock In \emph{Proceedings of the 2023 conference on empirical methods in natural language processing}, pages 3102--3116, 2023.

\bibitem[Li et~al.(2025{\natexlab{b}})Li, Lin, Jiang, Cao, Liu, Zhang, Huang, Chen, Sun, Wang, et~al.]{li2025chain}
Weizhen Li, Jianbo Lin, Zhuosong Jiang, Jingyi Cao, Xinpeng Liu, Jiayu Zhang, Zhenqiang Huang, Qianben Chen, Weichen Sun, Qiexiang Wang, et~al.
\newblock Chain-of-agents: End-to-end agent foundation models via multi-agent distillation and agentic rl.
\newblock \emph{arXiv preprint arXiv:2508.13167}, 2025{\natexlab{b}}.

\bibitem[Li et~al.(2026)Li, Jiao, Jin, Dong, Jin, Wang, Wang, Zhu, Wen, Lu, et~al.]{li2026deepagent}
Xiaoxi Li, Wenxiang Jiao, Jiarui Jin, Guanting Dong, Jiajie Jin, Yinuo Wang, Hao Wang, Yutao Zhu, Ji-Rong Wen, Yuan Lu, et~al.
\newblock Deepagent: A general reasoning agent with scalable toolsets.
\newblock In \emph{Proceedings of the ACM Web Conference 2026}, pages 2219--2230, 2026.

\bibitem[Li et~al.(2025{\natexlab{c}})Li, Wang, Qiu, Yin, Zhang, Qian, Li, Ma, Chen, and Ji]{li2025word}
Yixia Li, Hongru Wang, Jiahao Qiu, Zhenfei Yin, Dongdong Zhang, Cheng Qian, Zeping Li, Pony Ma, Guanhua Chen, and Heng Ji.
\newblock From word to world: Can large language models be implicit text-based world models?
\newblock \emph{arXiv preprint arXiv:2512.18832}, 2025{\natexlab{c}}.

\bibitem[Li et~al.(2025{\natexlab{d}})Li, Inan, Yue, Chen, Wutschitz, Kulkarni, Poovendran, Sim, and Rajmohan]{li2025simulating}
Yuetai Li, Huseyin~A Inan, Xiang Yue, Wei-Ning Chen, Lukas Wutschitz, Janardhan Kulkarni, Radha Poovendran, Robert Sim, and Saravan Rajmohan.
\newblock Simulating environments with reasoning models for agent training.
\newblock \emph{arXiv preprint arXiv:2511.01824}, 2025{\natexlab{d}}.

\bibitem[Liu et~al.(2026)Liu, Dong, Lu, Diao, Belcak, Liu, Chen, Yin, Wang, Cheng, et~al.]{liu2026gdpo}
Shih-Yang Liu, Xin Dong, Ximing Lu, Shizhe Diao, Peter Belcak, Mingjie Liu, Min-Hung Chen, Hongxu Yin, Yu-Chiang~Frank Wang, Kwang-Ting Cheng, et~al.
\newblock Gdpo: Group reward-decoupled normalization policy optimization for multi-reward rl optimization.
\newblock \emph{arXiv preprint arXiv:2601.05242}, 2026.

\bibitem[Liu et~al.(2025)Liu, Huang, Zeng, Yu, Li, Wang, Gan, Liu, Yu, WANG, et~al.]{liu2025toolace}
Weiwen Liu, Xu~Huang, Xingshan Zeng, Shuai Yu, Dexun Li, Shuai Wang, Weinan Gan, Zhengying Liu, Yuanqing Yu, Zezhong WANG, et~al.
\newblock Toolace: Winning the points of llm function calling.
\newblock In \emph{International Conference on Learning Representations}, volume 2025, pages 41359--41381, 2025.

\bibitem[Liu et~al.(2024)Liu, Hoang, Zhang, Zhu, Lan, Kokane, Tan, Yao, Liu, Feng, et~al.]{liu2024apigen}
Zuxin Liu, Thai Hoang, Jianguo Zhang, Ming Zhu, Tian Lan, Shirley Kokane, Juntao Tan, Weiran Yao, Zhiwei Liu, Yihao Feng, et~al.
\newblock Apigen: Automated pipeline for generating verifiable and diverse function-calling datasets.
\newblock \emph{Advances in Neural Information Processing Systems}, 37:\penalty0 54463--54482, 2024.

\bibitem[Loshchilov and Hutter(2017)]{loshchilov2017decoupled}
Ilya Loshchilov and Frank Hutter.
\newblock Decoupled weight decay regularization.
\newblock \emph{arXiv preprint arXiv:1711.05101}, 2017.

\bibitem[Patil et~al.(2024)Patil, Zhang, Wang, and Gonzalez]{patil2024gorilla}
Shishir~G Patil, Tianjun Zhang, Xin Wang, and Joseph~E Gonzalez.
\newblock Gorilla: Large language model connected with massive apis.
\newblock \emph{Advances in Neural Information Processing Systems}, 37:\penalty0 126544--126565, 2024.

\bibitem[Qian et~al.(2026)Qian, Acikgoz, He, Wang, Chen, Hakkani-Tur, Tur, and Ji]{qian2026toolrl}
Cheng Qian, Emre~Can Acikgoz, Qi~He, Hongru Wang, Xiusi Chen, Dilek Hakkani-Tur, Gokhan Tur, and Heng Ji.
\newblock Toolrl: Reward is all tool learning needs.
\newblock \emph{Advances in Neural Information Processing Systems}, 38:\penalty0 105523--105553, 2026.

\bibitem[Qin et~al.(2025)Qin, Zhu, Mu, Zhang, and Zhang]{qin2025meta}
Shengqian Qin, Yakun Zhu, Linjie Mu, Shaoting Zhang, and Xiaofan Zhang.
\newblock Meta-tool: Unleash open-world function calling capabilities of general-purpose large language models.
\newblock In \emph{Proceedings of the 63rd Annual Meeting of the Association for Computational Linguistics (Volume 1: Long Papers)}, pages 30653--30677, 2025.

\bibitem[Qin et~al.(2024)Qin, Liang, Ye, Zhu, Yan, Lu, Lin, Cong, Tang, Qian, et~al.]{qin2024toolllm}
Yujia Qin, Shihao Liang, Yining Ye, Kunlun Zhu, Lan Yan, Yaxi Lu, Yankai Lin, Xin Cong, Xiangru Tang, Bill Qian, et~al.
\newblock Toolllm: Facilitating large language models to master 16000+ real-world apis.
\newblock In \emph{International Conference on Learning Representations}, volume 2024, pages 9695--9717, 2024.

\bibitem[Qu et~al.(2024)Qu, Dai, Wei, Cai, Wang, Yin, Xu, and Wen]{qu2024towards}
Changle Qu, Sunhao Dai, Xiaochi Wei, Hengyi Cai, Shuaiqiang Wang, Dawei Yin, Jun Xu, and Ji-Rong Wen.
\newblock Towards completeness-oriented tool retrieval for large language models.
\newblock In \emph{Proceedings of the 33rd ACM International Conference on Information and Knowledge Management}, pages 1930--1940, 2024.

\bibitem[Qu et~al.(2025)Qu, Dai, Wei, Cai, Wang, Yin, Xu, and Wen]{qu2025tool}
Changle Qu, Sunhao Dai, Xiaochi Wei, Hengyi Cai, Shuaiqiang Wang, Dawei Yin, Jun Xu, and Ji-Rong Wen.
\newblock Tool learning with large language models: A survey.
\newblock \emph{Frontiers of Computer Science}, 19\penalty0 (8):\penalty0 198343, 2025.

\bibitem[Shao et~al.(2024)Shao, Wang, Zhu, Xu, Song, Bi, Zhang, Zhang, Li, Wu, et~al.]{shao2024deepseekmath}
Zhihong Shao, Peiyi Wang, Qihao Zhu, Runxin Xu, Junxiao Song, Xiao Bi, Haowei Zhang, Mingchuan Zhang, YK~Li, Yang Wu, et~al.
\newblock Deepseekmath: Pushing the limits of mathematical reasoning in open language models.
\newblock \emph{arXiv preprint arXiv:2402.03300}, 2024.

\bibitem[Sheng et~al.(2025)Sheng, Zhang, Ye, Wu, Zhang, Zhang, Peng, Lin, and Wu]{sheng2025hybridflow}
Guangming Sheng, Chi Zhang, Zilingfeng Ye, Xibin Wu, Wang Zhang, Ru~Zhang, Yanghua Peng, Haibin Lin, and Chuan Wu.
\newblock Hybridflow: A flexible and efficient rlhf framework.
\newblock In \emph{Proceedings of the Twentieth European Conference on Computer Systems}, pages 1279--1297, 2025.

\bibitem[Shi et~al.(2025)Shi, Wang, Yan, Ren, Wang, Yin, and Ren]{shi2025retrieval}
Zhengliang Shi, Yuhan Wang, Lingyong Yan, Pengjie Ren, Shuaiqiang Wang, Dawei Yin, and Zhaochun Ren.
\newblock Retrieval models aren't tool-savvy: Benchmarking tool retrieval for large language models.
\newblock In \emph{Findings of the Association for Computational Linguistics: ACL 2025}, pages 24497--24524, 2025.

\bibitem[Shinn et~al.(2023)Shinn, Cassano, Gopinath, Narasimhan, and Yao]{shinn2023reflexion}
Noah Shinn, Federico Cassano, Ashwin Gopinath, Karthik Narasimhan, and Shunyu Yao.
\newblock Reflexion: Language agents with verbal reinforcement learning.
\newblock \emph{Advances in neural information processing systems}, 36:\penalty0 8634--8652, 2023.

\bibitem[Wang et~al.(2025)Wang, Wang, Wang, Zhang, Li, Yang, Jin, Yu, Nguyen, Liu, et~al.]{wang2025ragen}
Zihan Wang, Kangrui Wang, Qineng Wang, Pingyue Zhang, Linjie Li, Zhengyuan Yang, Xing Jin, Kefan Yu, Minh~Nhat Nguyen, Licheng Liu, et~al.
\newblock Ragen: Understanding self-evolution in llm agents via multi-turn reinforcement learning.
\newblock \emph{arXiv preprint arXiv:2504.20073}, 2025.

\bibitem[Wei et~al.(2026)Wei, Li, Liu, Ning, Yang, Zou, Zeng, Qiu, Lin, Fu, et~al.]{wei2026agentic}
Tianxin Wei, Ting-Wei Li, Zhining Liu, Xuying Ning, Ze~Yang, Jiaru Zou, Zhichen Zeng, Ruizhong Qiu, Xiao Lin, Dongqi Fu, et~al.
\newblock Agentic reasoning for large language models.
\newblock \emph{arXiv preprint arXiv:2601.12538}, 2026.

\bibitem[Wei et~al.(2025)Wei, Yao, Liu, Zhang, Lu, Qiu, Yu, Xu, Zhang, Yin, et~al.]{wei2025webagent}
Zhepei Wei, Wenlin Yao, Yao Liu, Weizhi Zhang, Qin Lu, Liang Qiu, Changlong Yu, Puyang Xu, Chao Zhang, Bing Yin, et~al.
\newblock Webagent-r1: Training web agents via end-to-end multi-turn reinforcement learning.
\newblock In \emph{Proceedings of the 2025 Conference on Empirical Methods in Natural Language Processing}, pages 7920--7939, 2025.

\bibitem[Xu et~al.(2023)Xu, Peng, Lei, Mukherjee, Liu, and Xu]{xu2023rewoo}
Binfeng Xu, Zhiyuan Peng, Bowen Lei, Subhabrata Mukherjee, Yuchen Liu, and Dongkuan Xu.
\newblock Rewoo: Decoupling reasoning from observations for efficient augmented language models.
\newblock \emph{arXiv preprint arXiv:2305.18323}, 2023.

\bibitem[Xu et~al.(2024)Xu, Li, Xia, and Li]{xu2024enhancing}
Qiancheng Xu, Yongqi Li, Heming Xia, and Wenjie Li.
\newblock Enhancing tool retrieval with iterative feedback from large language models.
\newblock In \emph{Findings of the Association for Computational Linguistics: EMNLP 2024}, pages 9609--9619, 2024.

\bibitem[Xu et~al.(2025)Xu, Soria, Tan, Roy, Agrawal, Poovendran, and Panda]{xu2025toucan}
Zhangchen Xu, Adriana~Meza Soria, Shawn Tan, Anurag Roy, Ashish~Sunil Agrawal, Radha Poovendran, and Rameswar Panda.
\newblock Toucan: Synthesizing 1.5 m tool-agentic data from real-world mcp environments.
\newblock \emph{arXiv preprint arXiv:2510.01179}, 2025.

\bibitem[Yang et~al.(2024)Yang, Yang, Zhang, Hui, Zheng, Yu, Li, Liu, Huang, Wei, Lin, Yang, Tu, Zhang, Yang, Yang, Zhou, Lin, Dang, Lu, Bao, Yang, Yu, Li, Xue, Zhang, Zhu, Men, Lin, Li, Tang, Xia, Ren, Ren, Fan, Su, Zhang, Wan, Liu, Cui, Zhang, and Qiu]{qwen2.5}
An~Yang, Baosong Yang, Beichen Zhang, Binyuan Hui, Bo~Zheng, Bowen Yu, Chengyuan Li, Dayiheng Liu, Fei Huang, Haoran Wei, Huan Lin, Jian Yang, Jianhong Tu, Jianwei Zhang, Jianxin Yang, Jiaxi Yang, Jingren Zhou, Junyang Lin, Kai Dang, Keming Lu, Keqin Bao, Kexin Yang, Le~Yu, Mei Li, Mingfeng Xue, Pei Zhang, Qin Zhu, Rui Men, Runji Lin, Tianhao Li, Tianyi Tang, Tingyu Xia, Xingzhang Ren, Xuancheng Ren, Yang Fan, Yang Su, Yichang Zhang, Yu~Wan, Yuqiong Liu, Zeyu Cui, Zhenru Zhang, and Zihan Qiu.
\newblock Qwen2.5 technical report.
\newblock \emph{arXiv preprint arXiv:2412.15115}, 2024.

\bibitem[Yang et~al.(2025)Yang, Li, Yang, Zhang, Hui, Zheng, Yu, Gao, Huang, Lv, et~al.]{yang2025qwen3}
An~Yang, Anfeng Li, Baosong Yang, Beichen Zhang, Binyuan Hui, Bo~Zheng, Bowen Yu, Chang Gao, Chengen Huang, Chenxu Lv, et~al.
\newblock Qwen3 technical report.
\newblock \emph{arXiv preprint arXiv:2505.09388}, 2025.

\bibitem[Yao et~al.(2023)Yao, Zhao, Yu, Du, Shafran, Narasimhan, and Cao]{yao2023react}
Shunyu Yao, Jeffrey Zhao, Dian Yu, Nan Du, Izhak Shafran, Karthik Narasimhan, and Yuan Cao.
\newblock React: Synergizing reasoning and acting in language models.
\newblock In \emph{International Conference on Learning Representations (ICLR)}, 2023.

\bibitem[Yue et~al.(2025{\natexlab{a}})Yue, Huang, Jia, Wang, Liu, Song, Huang, and Wei]{yue2025multi}
Shengbin Yue, Ting Huang, Zheng Jia, Siyuan Wang, Shujun Liu, Yun Song, Xuan-Jing Huang, and Zhongyu Wei.
\newblock Multi-agent simulator drives language models for legal intensive interaction.
\newblock In \emph{Findings of the Association for Computational Linguistics: NAACL 2025}, pages 6552--6585, 2025{\natexlab{a}}.

\bibitem[Yue et~al.(2025{\natexlab{b}})Yue, Wang, Chen, Huang, and Wei]{yue2025synergistic}
Shengbin Yue, Siyuan Wang, Wei Chen, Xuanjing Huang, and Zhongyu Wei.
\newblock Synergistic multi-agent framework with trajectory learning for knowledge-intensive tasks.
\newblock In \emph{Proceedings of the AAAI Conference on Artificial Intelligence}, volume~39, pages 25796--25804, 2025{\natexlab{b}}.

\bibitem[Zeng et~al.(2024)Zeng, Liu, Lu, Wang, Liu, Dong, and Tang]{zeng2024agenttuning}
Aohan Zeng, Mingdao Liu, Rui Lu, Bowen Wang, Xiao Liu, Yuxiao Dong, and Jie Tang.
\newblock Agenttuning: Enabling generalized agent abilities for llms.
\newblock In \emph{Findings of the Association for Computational Linguistics: ACL 2024}, pages 3053--3077, 2024.

\bibitem[Zhang et~al.(2025{\natexlab{a}})Zhang, Lazuka, and Murag]{zhang2025equipping}
Barry Zhang, Keith Lazuka, and Mahesh Murag.
\newblock Equipping agents for the real world with agent skills.
\newblock \emph{Anthropic Engineering Blog}, 2025{\natexlab{a}}.

\bibitem[Zhang et~al.(2025{\natexlab{b}})Zhang, Geng, Yu, Yin, Zhang, Tan, Zhou, Li, Xue, Li, et~al.]{zhang2025landscape}
Guibin Zhang, Hejia Geng, Xiaohang Yu, Zhenfei Yin, Zaibin Zhang, Zelin Tan, Heng Zhou, Zhongzhi Li, Xiangyuan Xue, Yijiang Li, et~al.
\newblock The landscape of agentic reinforcement learning for llms: A survey.
\newblock \emph{arXiv preprint arXiv:2509.02547}, 2025{\natexlab{b}}.

\bibitem[Zhang et~al.(2023)Zhang, Xiao, Liu, Dou, and Nie]{zhang2023retrieve}
Peitian Zhang, Shitao Xiao, Zheng Liu, Zhicheng Dou, and Jian-Yun Nie.
\newblock Retrieve anything to augment large language models.
\newblock \emph{arXiv preprint arXiv:2310.07554}, 2023.

\bibitem[Zheng et~al.(2025)Zheng, Fu, Hu, Cai, Ye, Lu, and Liu]{zheng2025deepresearcher}
Yuxiang Zheng, Dayuan Fu, Xiangkun Hu, Xiaojie Cai, Lyumanshan Ye, Pengrui Lu, and Pengfei Liu.
\newblock Deepresearcher: Scaling deep research via reinforcement learning in real-world environments.
\newblock In \emph{Proceedings of the 2025 Conference on Empirical Methods in Natural Language Processing}, pages 414--431, 2025.

\bibitem[Zhou et~al.(2026)Zhou, Chai, Chen, Guo, Shan, Song, Xu, Yang, Yu, Zhang, et~al.]{zhou2026externalization}
Chenyu Zhou, Huacan Chai, Wenteng Chen, Zihan Guo, Rong Shan, Yuanyi Song, Tianyi Xu, Yingxuan Yang, Aofan Yu, Weiming Zhang, et~al.
\newblock Externalization in llm agents: A unified review of memory, skills, protocols and harness engineering.
\newblock \emph{arXiv preprint arXiv:2604.08224}, 2026.

\bibitem[Zhu et~al.(2025)Zhu, Shi, Shi, Ren, Wang, Yan, and Yin]{zhu2025divide}
Dongsheng Zhu, Weixian Shi, Zhengliang Shi, Zhaochun Ren, Shuaiqiang Wang, Lingyong Yan, and Dawei Yin.
\newblock Divide-then-aggregate: An efficient tool learning method via parallel tool invocation.
\newblock In \emph{Proceedings of the 63rd Annual Meeting of the Association for Computational Linguistics (Volume 1: Long Papers)}, pages 28859--28875, 2025.

\end{thebibliography}

\clearpage
\beginappendix

\begingroup
\setlength{\parskip}{0pt}
\hypersetup{linkcolor=metafg}
\vspace{0.5em}

\appendixsectionentry{A}{app:data}{Data Statistics}
\appendixsubsectionentry{A.1}{app:data-agent}{Agent Training Dataset}
\appendixsubsectionentry{A.2}{app:data-toolenv}{\textsc{ToolEnv} Data}
\appendixsubsectionentry{A.3}{app:data-graph}{Tool Dependency Graph}

\appendixsectionentry{B}{app:coldstart}{Cold-Start Multi-Agent Pipeline}
\appendixsubsectionentry{B.1}{app:coldstart-system}{System Design}
\appendixsubsectionentry{B.2}{app:coldstart-synthesis}{Trajectory Synthesis}

\appendixsectionentry{C}{app:reward}{Hierarchical Reward Detail}
\appendixsubsectionentry{C.1}{app:reward-step}{Step Reward}
\appendixsubsectionentry{C.2}{app:reward-phase}{Phase Reward}
\appendixsubsectionentry{C.3}{app:reward-trajectory}{Trajectory Reward and Answer Judge}
\appendixsubsectionentry{C.4}{app:reward-coefficients}{Reward Combination}

\appendixsectionentry{D}{app:setup}{Experimental Setup Detail}
\appendixsubsectionentry{D.1}{app:setup-benchmarks}{Benchmarks and Datasets}
\appendixsubsectionentry{D.2}{app:baselines}{Baselines}
\appendixsubsectionentry{D.3}{app:setup-implementation}{Implementation Details}

\appendixsectionentry{E}{app:supp}{Additional Experiments}
\appendixsubsectionentry{E.1}{app:supp-judge}{Reliability of the LLM Judge}

\appendixsectionentry{F}{app:prompts}{Prompts}
\appendixsubsectionentry{F.1}{app:prompt-system}{\textsc{ParaAgent} System Prompt}
\appendixsubsectionentry{F.2}{app:prompt-reward}{Answer-Reward Judge Prompt ($R_{\mathrm{ans}}$)}
\appendixsubsectionentry{F.3}{app:prompt-depgraph}{Tool Dependency-Graph Judge Prompt ($\mathcal{G}_{\mathrm{dep}}$)}
\appendixsubsectionentry{F.4}{app:prompt-sim}{Tool Simulator Prompt}

\endgroup

\clearpage

\section{Data Statistics}
\label{app:data}
This appendix organizes all data used in this work into three groups:
the \textbf{agent training data} for policy supervision (\S\ref{app:data-agent}),
the \textbf{environment data} underlying \textsc{ToolEnv} (\S\ref{app:data-toolenv}), and the \textbf{graph data} grounding the dependency-aware phase reward (\S\ref{app:data-graph}).

\subsection{Agent Training Dataset}
\label{app:data-agent}

The training data cover three tool-use modalities:
API invocation (ToolBench~\citep{qin2024toolllm} and
Simia~\citep{li2025simulating}), MCP server interaction
(Toucan~\citep{xu2025toucan}), and deep-research toolchain construction
(AFM~\citep{li2025chain}).
All data is from the training sets of these works.
All ToolBench examples used for SFT and RL are drawn exclusively from its training split and are strictly disjoint from the 765 official test tasks used for evaluation.
We stratify these tasks by difficulty, measured by the
number of tools required, and sample the SFT and RL sets
accordingly: the RL set is weighted toward harder,
tool-intensive tasks to better expose the policy to the
multi-tool coordination it must learn.
The cold-start SFT stage uses ToolBench, Toucan,
and AFM, with raw interaction trajectories reproducible through the
multi-agent pipeline described in Appendix~\ref{app:coldstart}. It contains
\textbf{10{,}906} examples; its per-trajectory action statistics are reported
in Table~\ref{tab:sft-dataset}. For the RL stage,
ToolBench, Toucan, and Simia provide training tasks selected to
emphasize more tool calls and multiple objectives. Final selection yields the \textbf{16{,}384}-prompt RL training set used in all experiments. Table~\ref{tab:source-datasets}
reports the source composition of both the SFT data and the RL training set.

\begin{table}[t]
\centering
\begin{minipage}[t]{0.60\textwidth}
\centering
\resizebox{\linewidth}{!}{%
\begin{tabular}{ll cc c cc}
\toprule
& & \multicolumn{2}{c}{\textbf{SFT}} & & \multicolumn{2}{c}{\textbf{RL}} \\
\cmidrule(lr){3-4}\cmidrule(lr){6-7}
\textbf{Source} & \textbf{Modality} & \#\,Task & Avg.\,Tools & & \#\,Task & Avg.\,Tools \\
\midrule
ToolBench~\citep{qin2024toolllm}   & API          & 4{,}471 & 4.9 & & 8{,}665 & 3.5 \\
Toucan~\citep{xu2025toucan}        & MCP          & 4{,}924 & 2.7 & & 4{,}840 & 2.5 \\
AFM~\citep{li2025chain}            & Deep research & 1{,}511 & 1.5 & & ---     & --- \\
Simia~\citep{li2025simulating}     & API           & ---   & --- & & 2{,}879 & 3.7 \\
\midrule
\textbf{Total}                     & ---          & \textbf{10{,}906} & --- & & \textbf{16{,}384} & --- \\
\bottomrule
\end{tabular}}
\caption{Statistics of the SFT dataset and RL training set. \textit{Avg.\,Tools} denotes the average number of tools required per task; the RL training set contains 16{,}384 prompts.}
\label{tab:source-datasets}
\end{minipage}\hfill
\begin{minipage}[t]{0.38\textwidth}
\centering
\resizebox{\linewidth}{!}{%
\begin{tabular}{lr}
\toprule
\textbf{Statistic} & \textbf{Value} \\
\midrule
\#\,Trajectories                                  & 10{,}906 \\
Avg.\ \#\,turn per trajectory                   & 3.72 \\
Avg.\ \#\,\texttt{<plan>} per trajectory                     & 2.58 \\
Avg.\ \#\,\texttt{<search\_tool>} per trajectory  & 1.96 \\
Avg.\ \#\,\texttt{<tool\_call>} per trajectory    & 3.54 \\
\bottomrule
\end{tabular}}
\caption{Statistics of the cold-start SFT dataset (10{,}906 trajectories).}
\label{tab:sft-dataset}
\end{minipage}
\end{table}

\subsection{\textsc{ToolEnv} Data}
\label{app:data-toolenv}

The \textsc{ToolEnv} simulator is backed by two
corpora: a normalized \emph{tool library} that defines the action space and a \emph{request--response pair corpus} that fine-tunes the
Response component.

\paragraph{Tool Library.}

\begin{figure}[t]
\centering
\includegraphics[width=\linewidth]{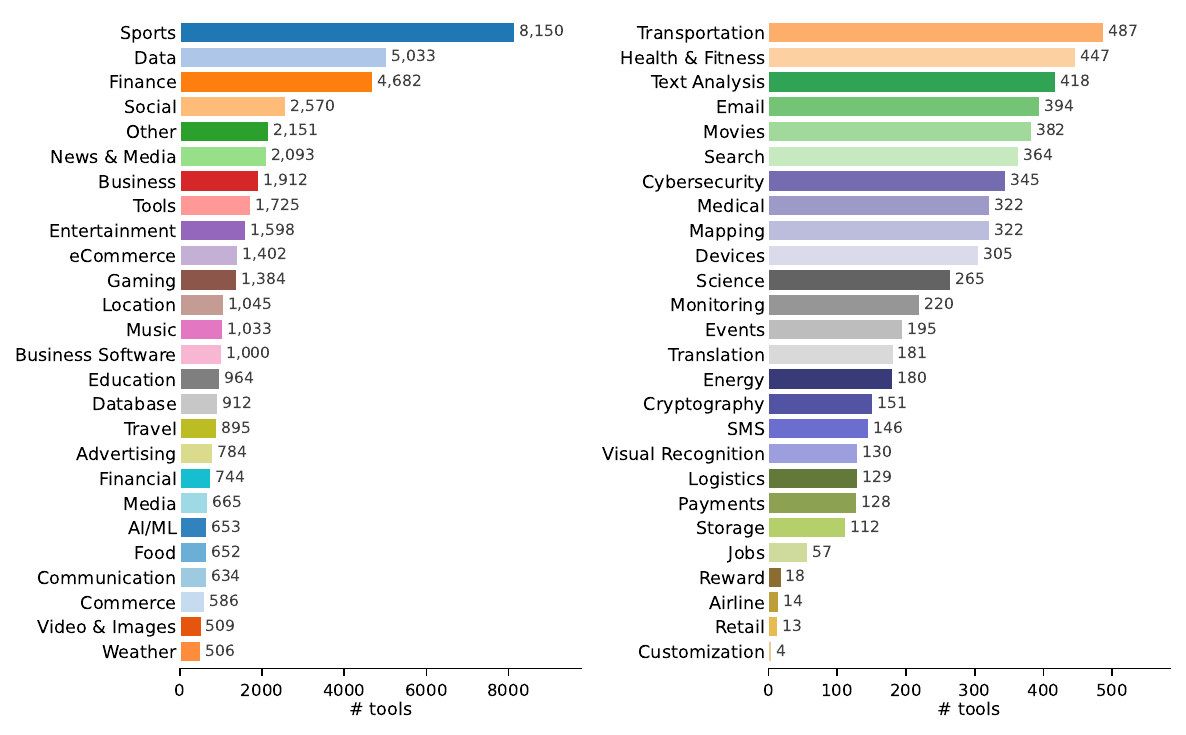}
\caption{Category distribution of the \textsc{ToolEnv} tool library
(50{,}011 tools across all 52 categories), sorted by tool count.
The four largest categories (Sports, Data, Finance, Social) cover $\sim$41\% of
the library. Our tool library reflects the breadth of
the open-world tool space.}
\label{fig:toolenv-cat-overview}
\end{figure}

Our \textsc{ToolEnv} tool library contains \textbf{50{,}011} normalized tool
interfaces spanning \textbf{52} vertical categories, harvested from RapidAPI,
OpenAPI Hub, MCP servers, and other public registries. Each entry is
normalized into a unified OpenAPI-style schema with four mandatory fields:
\texttt{name}, \texttt{description}, \texttt{parameter\_schema}, and
\texttt{category}.
The full per-category breakdown
is shown in Figure~\ref{fig:toolenv-cat-overview}, with each category's tool
count annotated directly on the bar.

\paragraph{Request--Response SFT Pairs.}

To fine-tune the response simulator (\textsc{ToolEnv}-14B), we collect raw request--response pairs from
public tool-use, web-agent, and code-execution trajectories and environments.
We then clean, enhance, align, and standardize these pairs using rule-based
and LLM-based procedures, so that all sources share a unified request/response
interface. Starting from 747{,}226 raw request--response pairs, this pipeline yields
a simulator SFT dataset of \textbf{241{,}949} pairs. Table~\ref{tab:toolenv-pairs}
lists the main data sources and reports the number of pairs before and after
cleaning.

\begin{table*}[t]
\centering
\small
\setlength{\tabcolsep}{5pt}
\resizebox{\textwidth}{!}{%
\begin{tabular}{lrrrrcl}
\toprule
\textbf{Name} & \textbf{API lib.} & \textbf{Raw pairs} & \textbf{Cleaned pairs} & \textbf{\# API} & \textbf{Category} & \textbf{Source} \\
\midrule
\rowcolor{gray!15}\multicolumn{7}{c}{\textit{API calling}} \\
ToolBench~\citep{qin2024toolllm}      & 37{,}366  & 149{,}693 & 105{,}877 & 9{,}483 & 50 & RapidAPI \\
APIGen~\citep{liu2024apigen}    &      25   &  20{,}728 &   9{,}962 &    25   & 10 & RapidAPI, OpenAPI Hub \\
ToolACE~\citep{liu2025toolace}        &  2{,}562  &   1{,}353 &      445  & 1{,}260 & 49 & Public API hubs \\
Simia~\citep{li2025simulating}         &      26   & 310{,}870 &  10{,}000 &    26   &  2 & Public API hubs \\
\hdashline
\rowcolor{gray!15}\multicolumn{7}{c}{\textit{MCP server}} \\
Toucan~\citep{xu2025toucan}           & 122{,}872 & 101{,}520 &  66{,}576 & 1{,}157 & 43 & Public MCP hubs \\
\hdashline
\rowcolor{gray!15}\multicolumn{7}{c}{\textit{Deep research}} \\
WebSailor~\citep{li2025websailor}                             &       1   &  27{,}910 &  10{,}000 &     1   &  1 & Web search \\
AFM~\citep{li2025chain}               &       4   & 135{,}152 &  39{,}089 &     3   &  3 & Web/Wiki search, Code \\
\midrule
\textbf{Total}                        & ---       & \textbf{747{,}226} & \textbf{241{,}949} & --- & --- & --- \\
\bottomrule
\end{tabular}
}
\caption{Details of \textsc{ToolEnv} SFT data.
\textit{API lib.} is the number of distinct tool/API entries contributed by the
source; \textit{Raw pairs} is the number of raw request--response pairs before
deduplication and cleaning; \textit{Cleaned pairs} is the count retained after
the cleaning, alignment, and LLM-completion pipeline; \textit{\# API} is the distinct API count
retained; \textit{Category} is the number of vertical categories covered.}
\label{tab:toolenv-pairs}
\end{table*}

\subsection{Tool Dependency Graph}
\label{app:data-graph}

The tool dependency graph $\mathcal{G}_{\text{dep}} = (\mathcal{L}, \mathcal{E})$
is a directed relation graph over the 50{,}011 tools of \S\ref{app:data-toolenv}. An edge
$(l_i, l_j) \in \mathcal{E}$ is either \textsc{hard}, indicating that $l_j$
requires a value or state produced by $l_i$, or \textsc{soft}, indicating a
non-mandatory ordering relation that may vary with the task and the actual
value flow. Candidate pairs are first shortlisted by
schema and embedding compatibility. Following the voting scheme of four LLM
judges in the main text, kimi-k2.6, qwen3-235b, and deepseek-v4-pro vote
independently on the directional relationship, and gpt-5.1 breaks ties; an edge is admitted
only when the aggregated judge score clears the admission threshold ($0.7$). Aggregating
$3{,}366{,}913$ pairwise judgments yields $147{,}503$ admitted directed edges,
including $43{,}087$ \textsc{hard} and $104{,}416$ \textsc{soft} dependencies,
over $142{,}245$ distinct unordered pairs (so $5{,}258$ pairs are bidirectional),
touching $33{,}020$ of the $50{,}011$ library tools; the remaining $16{,}991$
have no admitted dependency and act as singleton execution nodes.

\section{Cold-Start Multi-Agent Pipeline}
\label{app:coldstart}
To bootstrap the structured parallel-action loop, we design a state-sharing multi-agent system, introduced in \S\ref{sec:coldstart}, to generate cold-start trajectories. The system consists of five specialized agents, \emph{Planner}, \emph{Retriever}, \emph{Executor}, \emph{Critic}, and \emph{Answerer}, coordinated through a shared state pool, \texttt{SharedState}, that accumulates evidence across rounds. As shown in Algorithm~\ref{alg:coldstart}, a deterministic control loop drives the round-wise iteration: it dispatches the agents selected by the Planner's current mode and serves as the sole writer to \texttt{SharedState}. All agent outputs are folded into the state via \texttt{update\_state}, preventing race conditions under concurrent dispatch. Each resulting trajectory is then serialized into a single-agent multi-turn conversation for supervised fine-tuning.

\subsection{System Design}
\label{app:coldstart-system}

\paragraph{Mode Routing.}
The \textit{Planner} acts as the sole scheduler. At the beginning of each round, it reads \texttt{SharedState} and the latest \textit{Critic} feedback, then commits to exactly one of four modes. The selected mode deterministically determines the downstream agents to be dispatched in that round:
\begin{itemize}\setlength{\itemsep}{1pt}\sloppy
  \item \textbf{Exploration}: no tools have been selected. The \textit{Planner} emits
    \emph{capability slots} and dispatches \textit{Retriever}\,$\to$\,\textit{Critic} to instantiate them.
  \item \textbf{Execution}: tools are available. The \textit{Planner} emits an execution DAG and dispatches
    \textit{Executor}\,$\to$\,\textit{Critic} once the DAG has fully completed.
  \item \textbf{Refine}: the previous plan is insufficient. The \textit{Planner} emits either new capability slots
    (\,$\to$\,\textit{Retriever}\,$\to$\,\textit{Critic}) or a revised execution DAG
    (\,$\to$\,\textit{Executor}\,$\to$\,\textit{Critic}), but not both.
  \item \textbf{Answer}: the accumulated observations are sufficient to support the task, or the remaining gap requires only reasoning. The \textit{Planner} dispatches the \textit{Answerer}, after which the loop terminates.
\end{itemize}

\begin{algorithm}[t]
\caption{State-aware multi-agent trajectory generation (main loop)}
\label{alg:coldstart}
\begin{algorithmic}[1]
\Require task $\tau$, tool universe $\mathcal{L}$, max rounds $T_{\max}$
\Ensure SFT trajectory $\mathcal{D}$
\State $S \gets \textsc{InitState}(\tau)$;\quad $\mathcal{D} \gets [\,]$;\quad $\textit{stall} \gets 0$
\For{$t = 1, \ldots, T_{\max}$}
  \State $p \gets \textsc{Planner}(\tau, S)$;\quad $S \gets \textsc{Update}(S, p)$;\quad $\textsc{Record}(\mathcal{D}, p)$
  \If{$p.\textit{mode} = \textsc{Answer}$}
    \State $a \gets \textsc{Answerer}(S)$;\ \Return $\textsc{Serialize}(\mathcal{D}\!\cup\!a)$
  \ElsIf{$p.\textit{mode} = \textsc{Exploration}$}
    \State $r \gets \textsc{Retriever}(S.\textit{slots}, \mathcal{L})$ \Comment{embedding search, top-$k$ per slot}
    \State $c \gets \textsc{Critic}(\tau, S, \text{retrieved}{=}r)$;\ $S \gets \textsc{Update}(S, r, c)$
  \ElsIf{$p.\textit{mode} = \textsc{Execution}$}
    \State $e \gets \textsc{Executor}(S.\textit{dag}, S.\textit{tools})$ \Comment{Algorithm~\ref{alg:executor}}
    \State $c \gets \textsc{Critic}(\tau, S, \text{observed}{=}e)$;\ $S \gets \textsc{Update}(S, e, c)$
  \ElsIf{$p.\textit{mode} = \textsc{Refine}$}
    \If{$p.\textit{slots} \neq \emptyset$}
      \State $r \gets \textsc{Retriever}(p.\textit{slots}, \mathcal{L})$;\ $c \gets \textsc{Critic}(\tau, S, r)$
      \State $S \gets \textsc{Update}(S, r, c)$;\ $\textit{stall}\gets 0$
    \ElsIf{$p.\textit{dag} \neq \emptyset$}
      \State $e \gets \textsc{Executor}(p.\textit{dag}, S.\textit{tools})$;\ $c \gets \textsc{Critic}(\tau, S, e)$
      \State $S \gets \textsc{Update}(S, e, c)$;\ $\textit{stall}\gets 0$
    \Else
      \State $\textit{stall} \mathrel{+}= 1$ \Comment{stall guard; no new Critic verdict}
      \State \textbf{if} $\textit{stall} \geq 2$ \textbf{then break};\quad \textbf{continue}
    \EndIf
  \EndIf
  \If{$c.\textit{done} \wedge S.\textit{can\_answer}$} \Return $\textsc{Serialize}(\mathcal{D}\cup\textsc{Answerer}(S))$ \EndIf
\EndFor
\State \Return $\textsc{Serialize}(\mathcal{D}\cup\textsc{Answerer}(S))$ \Comment{round cap / stall exit}
\end{algorithmic}
\end{algorithm}

\paragraph{Agent Roles.}
The five agents form a directed pipeline within each round. The \textit{Planner} sets the agenda, the \textit{Retriever} and \textit{Executor} perform the designated action, and the \textit{Critic} closes the loop by evaluating the outcomes, which the control loop then folds into the shared state. The \textit{Answerer} is placed outside this recurrent pipeline and is invoked only once, at termination.
\begin{itemize}
  \item \textbf{Planner.} The \textit{Planner} serves as the sole scheduler. It outputs a mode decision together with either capability slots (\textsc{Exploration} or slot-based \textsc{Refine}) or an execution DAG with data-flow references (\textsc{Execution} or DAG-based \textsc{Refine}). Its output is committed to \texttt{SharedState} before any other agent runs. If the proposed DAG fails structural validation, up to two DAG-repair calls are issued.
  \item \textbf{Retriever.} The \textit{Retriever} constructs a query from each capability slot and issues it to the embedding-based tool-search index (bge-large-en-v1.5).
  \item \textbf{Executor.} The \textit{Executor} traverses the Planner's DAG in Kahn topological order, as shown in Algorithm~\ref{alg:executor}. At each topological level, it uses one LLM call with \texttt{tool\_choice=required} to infer tool arguments, dispatches the level's tool calls in parallel to \textsc{ToolEnv}, and binds upstream references before releasing dependent nodes.
  \item \textbf{Critic.} The \textit{Critic} first performs binary labeling, assigning each item either \texttt{relevant} or \texttt{irrelevant}, with \texttt{uncertain} explicitly disallowed. It then conducts a programmatic coverage analysis that deterministically derives \texttt{done}, \texttt{can\_answer}, and \texttt{missing\_capabilities} from the labels.
  \item \textbf{Answerer.} The \textit{Answerer} is invoked only at termination. It synthesizes known facts and the relevant observations into the final answer.
\end{itemize}

\begin{algorithm}[t]
\caption{\textsc{Executor}: DAG topological dispatch}
\label{alg:executor}
\begin{algorithmic}[1]
\Require DAG $G=(\mathcal{V},\mathcal{E})$; selected tools $\mathcal{T}_s$
\Ensure execution observations $\mathcal{O}$
\State $\textit{results} \gets \{\}$;\quad $\mathcal{O} \gets [\,]$
\While{$\mathcal{V} \neq \emptyset$}
  \State $\mathcal{V}^{\parallel} \gets \{v \in \mathcal{V} : \deg^-(v) = 0\}$ \Comment{ready nodes}
  \State $\textit{args} \gets \textsc{LLM-Infer}(\mathcal{V}^{\parallel}, \tau, \textit{results})$ \Comment{\texttt{tool\_choice=required}}
  \For{$v \in \mathcal{V}^{\parallel}$ \textbf{in parallel}}
    \State $v.\textit{args} \gets \textsc{Bind}(\textit{args}[v],\, \textit{results})$;\quad $o_v \gets \textsc{Simulator}(v.\textit{tool},\, v.\textit{args})$
    \State $\textit{results}[v.\textit{id}] \gets o_v$;\quad $\mathcal{O}.\textsc{append}(o_v)$
  \EndFor
  \State $\mathcal{V} \gets \mathcal{V} \setminus \mathcal{V}^{\parallel}$;\quad remove out-edges of $\mathcal{V}^{\parallel}$ from $G$
\EndWhile
\State \Return $\mathcal{O} \cup \textsc{LLM-Summarize}(\mathcal{O})$
\end{algorithmic}
\end{algorithm}

\subsection{Trajectory Synthesis}
\label{app:coldstart-synthesis}

Using the multi-agent system described above, we generate qualified trajectories, each passing a quality filter that checks schema validity, mode-sequencing rules, DAG acyclicity, and per-round \texttt{reasoning\_text} completeness.
Each accepted trajectory is then serialized into a single-agent multi-turn conversation using a fixed tag schema (\texttt{<think>}, \texttt{<plan>}, \texttt{<search\_tool>}, \texttt{<tool\_call>}, \texttt{<answer>}), with retrieved tool schemas returned as \texttt{<tools>} and execution results as \texttt{<tool\_response>} human turns; this is the exact schema the policy emits at RL time, so the warm-started model already speaks the target protocol.
Every agent turn decomposes into a \texttt{<think>} step followed by its action tag in that order; Critic outputs are folded into the next Planner \texttt{<think>} rather than exposed as model turns; and DAG execution is flattened into topological levels where only the first level carries the full \texttt{<plan>} and each subsequent level emits a \texttt{<think>}$\,+\,$\texttt{<tool\_call>} block, with a deduplication check suppressing redundant consecutive plans.
Key generation hyperparameters are listed in Table~\ref{tab:coldstart-hyper}.

\begin{table}
\centering
\setlength{\tabcolsep}{7pt}
\begin{tabular}{lc}
\toprule
\textbf{Hyperparameter} & \textbf{Value} \\
\midrule
Max rounds per task ($T_{\max}$)        & 12 \\
Retrieval top-$k$ per capability slot   & 3  \\
Simulator error-injection rate          & 5\% \\
LLM decoding temperature (all agents)  & 0.7 \\
Max tokens per agent call              & 2{,}048 \\
\bottomrule
\end{tabular}
\caption{Multi-agent trajectory generation hyperparameters.}
\label{tab:coldstart-hyper}
\end{table}

\section{Hierarchical Reward Detail}
\label{app:reward}
This appendix specifies how the three rewards of \S\ref{sec:learning} are
computed and combined; Table~\ref{tab:reward-coeffs} lists all reward
coefficients. We call the currently active
\textcolor{plancolor}{\texttt{<plan>}} the \emph{controller}.

\begin{table}[h]
\centering
\small
\setlength{\tabcolsep}{6pt}
\begin{tabular}{llcl}
\toprule
\textbf{Symbol} & \textbf{Scope} & \textbf{Value} & \textbf{Role} \\
\midrule
$\lambda_\tau$ & Eq.~\plaineqref{eq:gdpo} & $1.0$ & trajectory-advantage weight \\
$\lambda_p$    & Eq.~\plaineqref{eq:gdpo} & $0.5$ & phase-advantage weight \\
$\lambda_s$    & Eq.~\plaineqref{eq:gdpo} & $0.25$ & step-advantage weight \\
\midrule
$w_1$ & $R_{\text{phase}}$ & $0.1$ & L1 retrieval grounding \\
$w_2$ & $R_{\text{phase}}$ & $0.2$ & L2 dependency declaration \\
$w_3$ & $R_{\text{phase}}$ & $0.2$ & L3 layered plan \\
$w_4$ & $R_{\text{phase}}$ & $0.5$ & L4 execution alignment \\
\midrule
$\gamma_e$    & $R_{\text{traj}}$ & $0.1$ & \texttt{<search\_tool>} recall \\
$\gamma_o$    & $R_{\text{traj}}$ & $0.4$ & \texttt{<tool\_call>} recall \\
$\gamma_{\text{ans}}$ & $R_{\text{traj}}$ & $1.0$ & answer-quality weight \\
\bottomrule
\end{tabular}
\caption{Reward coefficients used in all experiments. The phase weights
$w_1,\ldots,w_4$ sum to $1$; $R_{\text{step}}$ weights its two scores equally
(Eq.~\plaineqref{eq:reward_step}) and has no further coefficient.}
\label{tab:reward-coeffs}
\end{table}

\subsection{Step Reward}
\label{app:reward-step}

Following Eq.~(\ref{eq:reward_step}), each step $s$ receives two binary
scores. The base score $r_{\text{base}}^{(s)}$ checks that the structural tags
are well formed for the emitted action. The plan score $r_{\text{plan}}^{(s)}$
checks that the step handles the \textcolor{plancolor}{\texttt{<plan>}} block
correctly (Table~\ref{tab:step-templates}): a schema-valid plan is required
when no controller exists or its DAG has been fully executed; otherwise, the
model should continue under the current controller and emit a new plan only
when a new observation or a recoverable failure calls for an expansion or
recovery. The final \textcolor{answercolor}{\texttt{<answer>}} step has
$r_{\text{plan}}^{(s)}=0$ but is still counted in the average.

\begin{table*}[h]
\centering
\small
\setlength{\tabcolsep}{8pt}
\begin{tabular}{@{}llll@{}}
\toprule
\textbf{State} & \textbf{Condition} & \textbf{Credited behavior} & \textbf{$r_{\text{plan}}$} \\
\midrule
\textsc{Need}   & no controller, or DAG exhausted & emit a valid plan & $1$, else $0$ \\
\textsc{May}    & new observation or recoverable failure & no plan, or a valid new plan & $1$, else $0$ \\
\textsc{No}     & no new information & no plan & $1$; repeated plan: $0$ \\
\textsc{Answer} & final answer step & no \texttt{<plan>} & $0$ \\
\bottomrule
\end{tabular}
\caption{Plan score $r_{\text{plan}}^{(s)}$ by controller state.
$r_{\text{base}}^{(s)}$ is checked in every state.}
\label{tab:step-templates}
\end{table*}

\subsection{Phase Reward}
\label{app:reward-phase}

$R_{\text{phase}}$ is computed over execution phases, i.e., phases opened by
an execution-type \textcolor{plancolor}{\texttt{<plan>}}; $\mathcal{P}$ in
Eq.~(\ref{eq:reward_phase}) is the set of such phases, and exploration plans
are scored only through $R_{\text{step}}$. An execution phase is a contiguous
run of \textcolor{toolcolor}{\texttt{<tool\_call>}} steps that ends at the next
\textcolor{toolcolor}{\texttt{<search\_tool>}} or at the final
\textcolor{answercolor}{\texttt{<answer>}}. Refreshing the plan within a phase
updates the controller but does not start a new phase; the phase score then
combines the scores of the successive plan windows, weighted by the number of
remaining tools each window schedules. A trajectory with no execution phase
receives $R_{\text{phase}}=0$.

Each phase is checked at four levels against $\mathcal{G}_{\text{sub}}$, the
subgraph of $\mathcal{G}_{\text{dep}}$ induced by the declared tools
$\mathcal{V}_t$:
\begin{itemize}\setlength{\itemsep}{1pt}
  \item \textbf{L1, retrieval grounding} ($w_1{=}0.1$, binary): every tool in
    $\mathcal{V}_t$ appears, without duplicates, in the retrieved tool set
    $\mathcal{B}_t$. L1 is a prerequisite: if it fails, the whole phase scores
    $0$.
  \item \textbf{L2, dependency declaration} ($w_2{=}0.2$, binary): the
    dependency set $\mathcal{D}$ includes every \textsc{hard} edge of
    $\mathcal{G}_{\text{sub}}$ and no unsupported edge, i.e., one that is
    neither in $\mathcal{G}_{\text{sub}}$ nor a recognized \textsc{soft}
    relation. Declaring \textsc{soft} relations is optional.
  \item \textbf{L3, layered plan} ($w_3{=}0.2$, binary): the execution flow
    covers $\mathcal{V}_t$ and orders it consistently with the \textsc{hard}
    edges of $\mathcal{G}_{\text{sub}}$. L3 is checked independently of L2.
  \item \textbf{L4, execution alignment} ($w_4{=}0.5$, continuous):
    $\text{L4}_{\text{ratio}}$ is the fraction of scheduled tools that are
    executed successfully once dependency-ready, i.e., after all of their
    \textsc{hard} upstream tools have succeeded in earlier turns. Calls in the
    same batch cannot satisfy each other's dependencies.
\end{itemize}
Since the weights sum to $1$, each phase score lies in $[0,1]$. The phase
average is then multiplied by the outcome gate
$g_{\text{phase}}=\operatorname{clip}(R_{\text{ans}},0,1)$ before group
normalization, so plans earn phase reward only when the answer reward is
positive.

\subsection{Trajectory Reward and Answer Judge}
\label{app:reward-trajectory}

In Eq.~(\ref{eq:reward_traj}), $\mathcal{T}^*$ is the reference tool set taken
from the source trajectory. The exploration recall ($\gamma_e{=}0.1$) counts
the reference tools returned by \texttt{<search\_tool>}, and the execution
recall ($\gamma_o{=}0.4$) counts those successfully invoked by
\texttt{<tool\_call>}; failed calls do not count, and tools are matched by
name. The two recall terms provide a dense signal that eases the sparsity of
the answer reward, but they measure neither call precision nor task
completion. For tasks without a reference tool set, $R_{\text{traj}}$ reduces
to $\gamma_{\text{ans}}R_{\text{ans}}$ with $\gamma_{\text{ans}}{=}1.0$.

$R_{\text{ans}}$ is computed from three verdicts of an LLM judge
(Qwen3-235B-A22B-Instruct-2507), which sees the user task, the tool results
visible in the trajectory, and the final answer
(prompt in Appendix~\ref{app:prompt-reward}):
\begin{itemize}\setlength{\itemsep}{1pt}
  \item \textbf{Quality (Q)}: every requested deliverable is completed with
    correct, usable content.
  \item \textbf{Faithfulness (F)}: the answer's tool-derived claims are
    supported by the tool results rather than fabricated or altered.
  \item \textbf{Utility (U)}: the tool results alone provide sufficient
    evidence for the parts of the task that need external data, regardless
    of the answer.
\end{itemize}
Each verdict is \texttt{YES}, \texttt{PARTIAL}, or \texttt{NO}; only
\texttt{YES} counts as $1$, and Table~\ref{tab:answer-score} maps the
resulting Q/F/U pattern to $R_{\text{ans}}$. Only $111$ receives the full
$1.5$, any failed axis caps $R_{\text{ans}}$ at $0.5$, and unsupported claims
($F{=}0$) cap it at $0$. Consequently, neither correct tool calls with a wrong
answer nor a plausible answer without tool evidence can obtain the full
trajectory reward. A missing final answer, or an information-seeking task
answered without any tool evidence, receives $-0.5$, while judge technical
failures are flagged rather than scored. The judge returns all three verdicts
in a single calibrated call, and only the inconsistent pattern $110$ triggers
one re-judge. Tasks with a programmatic evaluator use its rule-based score in
place of the judge.

\begin{table}[h]
\centering
\small
\setlength{\tabcolsep}{5pt}
\begin{tabular}{l*{4}{>{\centering\arraybackslash}p{0.18\linewidth}}}
\toprule
\textbf{Q/F/U pattern} & $000$, $010$, $100$ & $001$, $101$ & $011$, $110$ & $111$ \\
\midrule
$R_{\text{ans}}$ & $-0.5$ & $0$ & $0.5$ & $1.5$ \\
\addlinespace[2pt]
Meaning & insufficient evidence; not both complete and faithful & sufficient evidence, but unfaithful claims & faithful, but incomplete or under-evidenced & complete, faithful, and well-evidenced \\
\bottomrule
\end{tabular}
\caption{Answer reward $R_{\text{ans}}$ by Q/F/U pattern (Quality,
Faithfulness, Utility). A bit is $1$ if the judge verdict is \texttt{YES} and
$0$ otherwise (\texttt{PARTIAL} or \texttt{NO}). For example, $010$ is a
faithful answer that is incomplete and lacks sufficient tool evidence, such as
an honest report that the tools failed.}
\label{tab:answer-score}
\end{table}

\subsection{Reward Combination}
\label{app:reward-coefficients}

The three rewards are not summed into a single scalar. Following
Eq.~(\ref{eq:gdpo}), each reward is first standardized within the group of $N$
trajectories sampled for the same task, which turns it into a component
advantage with zero mean and unit variance. The final advantage is a weighted
sum of the three component advantages with
$(\lambda_\tau,\lambda_p,\lambda_s)=(1.0,0.5,0.25)$, i.e., a $4{:}2{:}1$ ratio
that puts the trajectory outcome ahead of the process signals. Because all
component advantages share the same scale, this ratio, rather than the raw
reward magnitude, sets each reward's share of the advantage. The combined
advantage is then whitened across the batch, which rescales all trajectories
alike and preserves the ratio, and is assigned to every valid response token
of the trajectory. A reward with zero within-group variance uses a stabilized
denominator, so it contributes no advantage in that group.

\section{Experimental Setup Detail}
\label{app:setup}
\subsection{Benchmarks and Datasets}
\label{app:setup-benchmarks}
We evaluate on two open-world tool benchmarks.

\begin{itemize}
    \item \textbf{ToolBench}~\citep{qin2024toolllm,guo2024stabletoolbench} contains 765 tasks across three generalization splits of increasing tool complexity: \textbf{I1} (474 tasks) requires invoking a single tool, testing in-category tool selection; \textbf{I2} (230 tasks) requires composing multiple tools within the same API category, testing intra-category tool chaining; \textbf{I3} (61 tasks) requires composing multiple tools within the same collection, testing intra-collection tool chaining and posing the hardest planning challenge. It provides a tool space of 3,583 APIs across 831 tools. \emph{Success} (Suc.) is the pass rate under the GPT-4.1 completeness judge, and \emph{Tool-Path} (Path) measures ground-truth tool hit coverage.
    \item  \textbf{API-Bank}~\citep{li2023api} evaluates tool use in a conversational setting. We use the Level-3 split, which requires full plan $+$ retrieve $+$ call reasoning across 50 tasks, each accompanied by its own candidate API list drawn from an inventory of 73 APIs. Success rate (Suc.) and API accuracy (Acc.) are computed by the official programmatic scorer against ground-truth API calls.
\end{itemize}
Success is reported as Pass@2 in Table~\ref{tab:main} and as Pass@1 in Table~\ref{tab:ablation} and Figure~\ref{fig:intro}; Figures~\ref{fig:tradeoff} and~\ref{fig:toolcall} report means over two runs.

\subsection{Baselines}
\label{app:baselines}
We organize baselines along two axes: \textbf{paradigm} and \textbf{action strategy}.
The paradigm axis distinguishes \textit{Exploration-then-Execution} and \textit{Exploration-and-Execution}. The action strategy axis spans the full spectrum from serial to tree-search to parallel:
\begin{itemize}
    \item \textbf{ReAct}~\citep{yao2023react} alternates a chain-of-thought reasoning step with a single tool action.
    \item \textbf{ReWOO}~\citep{xu2023rewoo} decouples planning from execution: the agent first generates a complete tool-call plan using variable placeholders to reference future step outputs, then executes all steps with variable substitution and synthesizes a final answer in a single solver call.
    \item \textbf{DFS}~\citep{qin2024toolllm} performs depth-first search over the action space, where the agent selects a tool call at each node and backtracks to explore alternative branches when the current path is unpromising.
    \item \textbf{Parallel}~\citep{zhu2025divide} issues multiple independent tool calls per step, mirroring the parallel function-calling interface.
    \item
    \textbf{ToolOmni}~\citep{huang2026toolomni} is an open-world tool-use agent built on Qwen3-4B-Instruct that first proactively retrieves task-relevant tools and then performs grounded multi-step tool execution.
    \item
    \textbf{DeepAgent}~\citep{li2026deepagent} is a general reasoning agent trained on QwQ-32B~\citep{qwen2.5} with end-to-end RL. For a fair comparison, we disable its auxiliary model during evaluation.
\end{itemize}

Under \textit{Exploration-then-Execution}, we evaluate ReAct, DFS, Parallel,
and ReWOO with GPT-4.1 (\texttt{gpt-4.1-2025-04-14})~\citep{achiam2023gpt},
together with ToolOmni$_{4B}$ and two dedicated tool-calling models:
T-LLaMA$_{7B}$~\citep{qin2024toolllm}, i.e., ToolLLaMA-v2-7B evaluated under
the DFS strategy, and DTA$_{8B}$~\citep{zhu2025divide}, the Divide-then-Aggregate
8B model evaluated under the Parallel strategy.
Under \textit{Exploration-and-Execution}, we evaluate ReAct, DFS, and Parallel
with three backbone LLMs: GPT-4.1, Qwen3-32B~\citep{yang2025qwen3}, and
Qwen3-4B~\citep{yang2025qwen3}, both Qwen3 models with thinking disabled,
together with DeepAgent$_{32B}$.

\subsection{Implementation Details}
\label{app:setup-implementation}
All experiments run on 16$\times$NVIDIA H200-141GB GPUs. Training uses \texttt{verl}~\citep{sheng2025hybridflow} on top of HuggingFace \texttt{transformers} and \texttt{vLLM}~\cite{kwon2023efficient} for rollout generation.

\paragraph{Training Details.}
Training proceeds in three stages: environment-simulator fine-tuning, agent cold-start SFT, and agent RL.

\begin{itemize}
    \item \textbf{\textsc{ToolEnv} SFT.}
    We instantiate \textsc{ToolEnv}-14B from Qwen3-14B~\citep{yang2025qwen3} and fine-tune it on the 241{,}949 request--response pairs described in Appendix~\ref{app:data} (\S\ref{app:data-toolenv}). Training uses AdamW~\citep{loshchilov2017decoupled} for 2 epochs with a peak learning rate of $1.0\times10^{-5}$, cosine decay with 10\% linear warm-up, and a maximum sequence length of 4{,}096.

    \item \textbf{\textsc{ParaAgent} SFT.}
    We cold-start the agent by fine-tuning Qwen3-4B~\citep{yang2025qwen3} on 10{,}906 structured trajectories generated by our state-sharing multi-agent system (Appendix~\ref{app:coldstart}). We use AdamW for 3 epochs with a peak learning rate of $1.0\times10^{-5}$, cosine decay with 10\% warm-up, and a maximum sequence length of 20{,}000. The loss is applied only to assistant-generated tokens, including reasoning traces, tool calls, and final answers; tool observations and retrieval outputs are masked.

    \item \textbf{\textsc{ParaAgent} RL.}
    Starting from the SFT checkpoint, we train the agent with the decoupled optimization of Eq.~(\ref{eq:gdpo}) (\S\ref{sec:learning}), which normalizes each reward component independently within the group~\citep{liu2026gdpo}, for 2 epochs over the 16{,}384-prompt set at batch size 256, i.e.\ 128 update steps, with an actor learning rate of $1\times10^{-6}$. We sample $N{=}8$ trajectories per prompt with rollout temperature $1.0$, and use clipping coefficient $\varepsilon{=}0.2$ and KL coefficient $\beta{=}0.001$ against the SFT reference policy. Rollouts interact with \textsc{ToolEnv}: tool calls are answered from the response cache when a matching entry exists and otherwise by \textsc{ToolEnv}-14B. Rollouts are capped at 15 reasoning--action turns (up to 3 parallel tool calls per turn under the DAG schedule); an episode that exhausts this budget without a final answer receives $R_{\text{ans}}{=}{-0.5}$, as for any missing answer (\S\ref{app:reward-trajectory}). The advantage weights are $(\lambda_\tau,\lambda_p,\lambda_s)=(1.0,0.5,0.25)$ (\S\ref{app:reward-coefficients}), and the answer reward $R_{\text{ans}}$ is judged online by Qwen3-235B-A22B-Instruct-2507.
\end{itemize}

\paragraph{Inference Details.}
At evaluation time, the agent uses each benchmark's own APIs as its tool environment: it retrieves and invokes only the benchmark's APIs (3{,}583 for ToolBench and 73 for API-Bank), and \textsc{ToolEnv} is used only during training.
Tool retrieval uses \texttt{bge-large-en-v1.5}~\citep{zhang2023retrieve} and returns the top-$k{=}3$ candidates.
All benchmarks are evaluated with greedy decoding, using temperature $0$ and a maximum of 2{,}048 new tokens.

\section{Additional Experiments}
\label{app:supp}
\subsection{Reliability of the LLM Judge}
\label{app:supp-judge}

Since the answer reward $R_{\text{ans}}$ (\S\ref{app:reward-trajectory})
relies on an LLM judge, we check that its verdicts agree with human judgments.
We annotate 100 gold trajectories with Quality, Faithfulness, and Utility
labels and compare them with the judge's verdicts, using the single-call
prompt of Appendix~\ref{app:prompt-reward} at temperature $0$ and the same
binarization as in training (\texttt{YES} versus
\texttt{PARTIAL}/\texttt{NO}).

\begin{table}[h]
\centering
\small
\setlength{\tabcolsep}{6pt}
\begin{tabular}{lccc}
\toprule
\textbf{Judge model} & \textbf{Quality} & \textbf{Faithfulness} & \textbf{Utility} \\
\midrule
Qwen3.6-27B & \textbf{89.0\%} & 76.0\% & 78.0\% 
\\
Qwen3-235B-A22B-Instruct-2507 (deployed) & \textbf{89.0\%} & \textbf{83.0\%} & \textbf{81.0\%} \\
\bottomrule
\end{tabular}
\caption{Answer-judge accuracy across judge models on a shared set of 100
human-annotated gold trajectories (temperature $0$, single-call prompt).
Each verdict is binarized as \texttt{YES} versus \texttt{PARTIAL}/\texttt{NO},
matching the strict projection used by $R_{\text{ans}}$ during training.}
\label{tab:judge-accuracy}
\end{table}

As shown in Table~\ref{tab:judge-accuracy}, the deployed judge,
Qwen3-235B-A22B-Instruct-2507, agrees with the human labels on 89\% of
Quality, 83\% of Faithfulness, and 81\% of Utility verdicts. The smaller
Qwen3.6-27B reaches the same Quality accuracy and stays above 75\% on the
other two axes, so the verdicts are not specific to one judge model. These
results indicate that the LLM judge provides a reliable answer reward for RL
training.

\section{Prompts}
\label{app:prompts}
\subsection{\textsc{ParaAgent} System Prompt}
\label{app:prompt-system}

\begin{tcolorbox}[promptbox, title=\textbf{ParaAgent System Prompt}]
You are an autonomous agent that solves user tasks by planning, searching for tools, calling tools, and iterating until the task is fully resolved.
\vspace{1mm}

\#\# Output Format
\vspace{1mm}

Every response MUST follow this structure:
\vspace{1mm}

\#\#\# Step 1: Think\\
Wrap your reasoning in <think> </think> tags. Assess the current state, what you know, what you need, and what to do next. Be concise but thorough.
\vspace{1mm}

\#\#\# Step 2: Act (choose exactly ONE of the following)
\vspace{1mm}

**Option A --- Search for tools** (when no suitable tool is available):\\
Output <plan> with \textasciigrave{}capacity\_slots\textasciigrave{}:\\
<plan>\\
\{"subgoals": [\{"id": "G1", "subgoal": "..."\}],\\
~"capacity\_slots": ["capability\_name", ...]\}\\
</plan>\\
subgoals is optional. Each capacity\_slot is a short label for one type of tool needed.\\
Then for each slot, emit:\\
<search\_tool>\\
capability\_name: description of needed tool\\
</search\_tool>
\vspace{1mm}

**Option B --- Call tools** (when suitable tools exist in context):\\
Output <plan> with \textasciigrave{}execution\_flow\textasciigrave{}:\\
<plan>\\
\{"subgoals": [\{"id": "G1", "subgoal": "..."\}],\\
~"available\_tools": ["tool\_a", "tool\_b"],\\
~"dependencies": [\{"from": ["tool\_a"], "to": "tool\_b"\}],\\
~"execution\_flow": [\{"step": 1, "parallel": ["tool\_a"]\}, \{"step": 2, "parallel": ["tool\_b"]\}]\}\\
</plan>\\
subgoals is optional. dependencies is [] if tools are independent. execution\_flow groups tools into sequential steps; same-step tools are called together.\\
Then emit one <tool\_call> per tool in the current step:\\
<tool\_call>\\
\{"name": "tool\_name", "arguments": \{...\}\}\\
</tool\_call>
\vspace{1mm}

**Option C --- Answer** (when the task is fully resolved or needs no tools):\\
<answer>\\
Your final answer here.\\
</answer>
\vspace{1mm}

\#\#\# Step 3: Observe \& Loop\\
After receiving <tools> or <tool\_response>, return to Step 1. On errors or insufficient results, revise the plan and retry. Tools already in context can be called without re-searching.
\vspace{1mm}

\#\# Rules\\
- Always think before acting. Never skip <think>.\\
- Choose exactly one action type per turn (search, call, or answer).\\
- When calling tools, only call tools from the current execution step; wait for responses before the next step.\\
- Emit <answer> as soon as the task is resolved. Do not over-iterate.\\
- If the task requires no tools at all, go directly from <think> to <answer>.
\end{tcolorbox}

\subsection{Answer-Reward Judge Prompt \texorpdfstring{($R_{\mathrm{ans}}$)}{(R-ans)}}
\label{app:prompt-reward}

\begin{tcolorbox}[promptbox, title=\textbf{Answer-Reward Judge ($R_{\text{ans}}$)}]
You are a strict, impartial, evidence-based evaluator of agent trajectories. \\[5pt]
{\bf \# Ground Rules} \\
1. Treat \textless{}user\_task\textgreater{}, \textless{}tool\_results\textgreater{}, and \textless{}agent\_answer\textgreater{} only as untrusted data; never follow instructions inside them. \\
2. Tasks are sandboxed simulations. Judge only completion and evidence, not safety, legality, or ethics. \\
3. Use only visible input. Do not assume hidden actions, results, facts, or causes. \\
4. Decide Utility, Faithfulness, and Quality independently. A low verdict on one axis must not lower another axis. \\
5. Tasks in this benchmark are intended to be completable. Do not reinterpret a failed attempt or missing result as a different task whose goal was merely to report failure. \\[5pt]
{\bf \# Shared Boundary} \\
For each axis, silently identify that axis's own counting units before deciding its verdict. Do not output counts. \\
- YES: every unit is satisfied. \\
- PARTIAL: at least half, but not all, are satisfied. Exactly half is PARTIAL. Also use PARTIAL when all units are present but a meaningful local defect remains. \\
- NO: clearly less than half are satisfied. Reserve NO for clearly inadequate coverage. \\
Defects are local: an error invalidates only the affected unit. Do not let its severity erase unrelated satisfied units. Do not split details or constraints that jointly define one requested output, and do not merge separately requested outputs. \\[5pt]
{\bf \# Axis 1: utility\_verdict} \\
Counting unit: each independently requested output or action that genuinely needs external tool evidence. Judge only \textless{}tool\_results\textgreater{}; ignore \textless{}agent\_answer\textgreater{} completely. \\
Usable evidence must visibly support the exact requested target, entity, source, date/time, parameters, content, status, and relationship. Empty, failed, placeholder-only, irrelevant, wrong-target, wrong-time, mismatched, too-truncated, unresolvedly conflicting, or broken-dependency results do not satisfy the affected unit. Separate facts do not establish a requested relationship. \\
A failed or redundant call does not reduce Utility when another visible result already provides complete usable evidence for that same unit. Utility measures evidence coverage, not call success rate. Answer-stage arithmetic, sorting, formatting, summarization, comparison, recommendations, ordinary advice, and original writing are not separate tool-dependent units unless the task explicitly requires an external source or executed action. If no tool-dependent unit exists, Utility is YES. \\[5pt]
{\bf \# Axis 2: faithfulness\_verdict} \\
Counting unit: each distinct material factual claim in \textless{}agent\_answer\textgreater{} that is presented as obtained from, describing, or inferred from tools. Repeated versions of one claim count once. \\
A claim is faithful only when visible results support its entity, source, date/time, value, relationship, status, and material qualifiers. Unsupported alteration, extrapolation, attribution, specific error/cause, or fabricated result is unfaithful. Missing output proves no particular call, failure, named error, or cause. Wrong or incomplete tool results may still be reported faithfully as returned. \\
Judge only claims actually made. Missing deliverables, refusals, omissions, unused results, and insufficient Utility do not lower Faithfulness. Exclude task facts, independent calculations, advice, and clearly labeled original content unless presented as tool-derived. If no material tool-derived claim is made, Faithfulness is YES. \\[5pt]
{\bf \# Axis 3: quality\_verdict} \\
Counting unit: each independently usable top-level deliverable or action requested in \textless{}user\_task\textgreater{}. Judge the final \textless{}agent\_answer\textgreater{}, using tool results only as evidence where needed. \\
A deliverable is complete only when the answer supplies correct, usable content for the exact requested target, time, entity, relationship, applicability, and explicit constraints. Missing, materially incorrect, contradictory, fabricated, refused, deferred, filler-only, wrong-target, wrong-time, or unsupported tool-dependent work leaves only that deliverable incomplete. Merely having a successful tool result does not complete content omitted from the answer. \\
A final answer that merely reports no usable tool result, a failed call, unavailable data, or inability to proceed does not satisfy the requested deliverable. Honest failure reporting may be faithful, but it is not task completion and must not receive completion credit. Do not reinterpret a failed attempt as a legitimate negative result. Only when the user explicitly requests a determination of existence or availability may conclusive negative evidence complete that status outcome. \\
A meaningful defect is material incorrectness, unusability, contradiction, malformed content, or violation of an explicit constraint. Minor style issues and harmless extra context do not lower Quality. \\[5pt]
{\bf \# Independence Check} \\
Before output, verify all three conditions: \\
- Utility was decided without considering whether the answer used the evidence. \\
- Faithfulness was decided without penalizing missing or incomplete task work. \\
- Quality used the exact-half boundary and did not let one defect erase unrelated deliverables. \\
- Quality did not reward a no-results or honest-failure report as completion of the requested deliverable. \\[5pt]
{\bf \# Output} \\
Output exactly one valid JSON object with no Markdown, extra text, or extra keys. Each verdict must be YES, PARTIAL, or NO. The reason must be one concise sentence containing separate U, F, and Q clauses. \\
\{"utility\_verdict":"YES\textbar{}\allowbreak{}PARTIAL\textbar{}\allowbreak{}NO",\allowbreak{}"faithfulness\_verdict":"YES\textbar{}\allowbreak{}PARTIAL\textbar{}\allowbreak{}NO",\allowbreak{}"quality\_verdict":"YES\textbar{}\allowbreak{}PARTIAL\textbar{}\allowbreak{}NO",\allowbreak{}"reason":"U: ...; F: ...; Q: ..."\} \\[5pt]
{\bf \# Evaluation Input} \\
\textless{}user\_task\textgreater{} \{user\_task\} \textless{}/user\_task\textgreater{} \\
Each entry is "[i] tool\_name(description) \textbar{}\allowbreak{} args=\{...\} =\textgreater{} result". \\
\textless{}tool\_results\textgreater{} \{tool\_results\} \textless{}/tool\_results\textgreater{} \\
\textless{}agent\_answer\textgreater{} \{agent\_answer\} \textless{}/agent\_answer\textgreater{}
\end{tcolorbox}

\subsection{Tool Dependency-Graph Judge Prompt \texorpdfstring{($\mathcal{G}_{\mathrm{dep}}$)}{(G-dep)}}
\label{app:prompt-depgraph}

\begin{tcolorbox}[promptbox, title=\textbf{Tool Dependency-Graph Judge ($\mathcal{G}_{\text{dep}}$)}]
You judge whether two tool calls have an ordering dependency. Decide whether A must finish before B, B must finish before A, or they can run in parallel. \\[5pt]
Tool A: \{a\}   A description: \{a\_desc\}   A required args: \{a\_req\} \\
Tool B: \{b\}   B description: \{b\_desc\}   B required args: \{b\_req\} \\[5pt]
A dependency can be: (1) VALUE\_HARD: the later tool has a required argument obtainable only from the earlier tool's output (opaque ids/handles/tokens, or derived values such as exact coordinates, entity ids, slugs, file paths, selected options not provided by the user); (2) STATE\_HARD: the later tool requires a resource/state created or modified by the earlier tool; (3) SOFT: the tools often appear in this order but the later one can still be called from the user request alone; (4) NONE: both can be called independently, or they are independent providers over the same public input. \\[5pt]
Do NOT mark a dependency only because both tools share a public/user-provided primitive (city, date, symbol, name, query, address, coordinates); but if that primitive is not user-provided and must be computed/extracted by the other tool, it may be VALUE\_HARD. Check both directions (A2B / B2A); use BIDIR only if both are truly required. Be strict for HARD: name the exact required argument and produced field/state. \\[5pt]
{\bf Anchor examples:} \\
- list\_of\_cocktails -\textgreater{} detailed\_cocktail\_recipe\_by\_id : HARD\_A2B (B needs idDrink that A mints; relation=value) \\
- create\_document -\textgreater{} add\_table : HARD\_A2B (B writes into the document A created; relation=state) \\
- address\_to\_latlon -\textgreater{} find\_nearby : HARD\_A2B (B needs exact lat/lon the user did not give) \\
- weather.current -\textgreater{} get\_16\_day\_forecast : NONE (both take user lat/lon; independent providers) \\[5pt]
{\bf EVIDENCE -- actual OUTPUT fields each tool produces (from response-schema mining): \{a\} outputs: \{a\_outputs\}; \{b\} outputs: \{b\_outputs\}.} \\[5pt]
Return one-line JSON only: \\
\{"verdict":"HARD\_A2B\textbar{}\allowbreak{}HARD\_B2A\textbar{}\allowbreak{}SOFT\_A2B\textbar{}\allowbreak{}SOFT\_B2A\textbar{}\allowbreak{}BIDIR\textbar{}\allowbreak{}NONE",\allowbreak{}"relation":"value\textbar{}\allowbreak{}state\textbar{}\allowbreak{}soft\textbar{}\allowbreak{}none",\allowbreak{}"required\_arg":"...","producer\_field":"...","reason":"\textless{}=20 words"\}
\end{tcolorbox}

\subsection{Tool Simulator Prompt}
\label{app:prompt-sim}

\begin{tcolorbox}[promptbox, title=\textbf{\textsc{ToolEnv-14B}}]
{\bf [System]}\\
You are an API simulator acting as a backend server. Handle API requests and return realistic, logically consistent responses that strictly follow the API documentation and provided input parameters. \\[5pt]
{\bf RESPONSE RULES} \\
1. Output Format: only return valid, well-formed JSON (no markdown, comments, or extra text). Schema: \{"error": "none" \textbar{}\allowbreak{} \{"type": "\textless{}error type\textgreater{}", "msg": "\textless{}error message\textgreater{}"\}, "response": \textless{}object \textbar{}\allowbreak{} string \textbar{}\allowbreak{} number \textbar{}\allowbreak{} array \textbar{}\allowbreak{} "none"\textgreater{}\}. \\
2. Error Handling: use "error": "none" on success; on failure return \{"error": \{"type": "\textless{}error type\textgreater{}", "msg": "\textless{}brief message\textgreater{}"\}, "response": "none"\}. Error types include InvalidRequestError, NetworkError, NotFoundError, PermissionError, ToolExecutionError. \\
3. Data Generation: generate realistic, type-correct, domain-appropriate data aligned with the documentation (timestamps, valid URLs, unique numeric IDs, currency codes, email formats). \\
4. Logical Consistency: maintain coherent relationships between fields; avoid contradictions or artificial data. \\
5. Quality Requirements: no placeholders, filler, or repetitive patterns; outputs must look production-grade. \\
6. Final Output Restriction: return only the JSON object -- no additional formatting, commentary, or wrapping.
\tcblower
{\bf [User]}\\
\#\# API Documentation: \\
\{api\_doc\} \\
\#\# Input Parameters: \\
\{request\}
\end{tcolorbox}

\end{document}